\documentclass[10pt,twocolumn,letterpaper]{article}

\usepackage{wacv}
\usepackage{times}
\usepackage{epsfig}
\usepackage{graphicx}
\usepackage{amsmath}
\usepackage{amssymb}

\usepackage{subcaption}
\usepackage{multirow}
\usepackage{float}
\usepackage[utf8x]{inputenc}
\usepackage{layouts}

\newcommand{\warp}{\textrm{warp}}

\newcommand{\thresh}{\textrm{thresh}}

\def\etc{\emph{etc}\onedot}

\newcommand{\textubf}[1]{\underline{\textbf{#1}}}

\def\wacvPaperID{654} 

\wacvfinalcopy 

\ifwacvfinal
\fi

\ifwacvfinal
\usepackage[breaklinks=true,bookmarks=false]{hyperref}
\else
\usepackage[pagebackref=true,breaklinks=true,colorlinks,bookmarks=false]{hyperref}
\fi

\begin{document}

\title{HyperCon: Image-To-Video Model Transfer for\\Video-To-Video Translation Tasks}

\author{Ryan Szeto$^\dagger$ \quad Mostafa El-Khamy$^\mathsection$ \quad Jungwon Lee$^\mathsection$ \quad Jason J. Corso$^\dagger$\\
$^\dagger$University of Michigan \quad $^\mathsection$Samsung Semiconductor, Inc.\\
{\tt\small \{szetor,jjcorso\}@umich.edu, \{mostafa.e,jungwon2.lee\}@samsung.com}
}

\maketitle
\thispagestyle{empty}

\begin{abstract}
  Video-to-video translation is more difficult than image-to-image translation due to the temporal consistency problem that, if unaddressed, leads to distracting flickering effects. Although video models designed from scratch produce temporally consistent results, training them to match the vast visual knowledge captured by image models requires an intractable number of videos. To combine the benefits of image and video models, we propose an image-to-video model transfer method called Hyperconsistency (HyperCon) that transforms any well-trained image model into a temporally consistent video model without fine-tuning. HyperCon works by translating a temporally interpolated video frame-wise and then aggregating over temporally localized windows on the interpolated video. It handles both masked and unmasked inputs, enabling support for even more video-to-video translation tasks than prior image-to-video model transfer techniques. We demonstrate HyperCon on video style transfer and inpainting, where it performs favorably compared to prior state-of-the-art methods without training on a single stylized or incomplete video. Our project website is available at \href{https://ryanszeto.com/projects/hypercon}{\texttt{ryanszeto.com/projects/hypercon}}.
\end{abstract}

\begin{figure*}
  \centering
  \includegraphics[width=\linewidth]{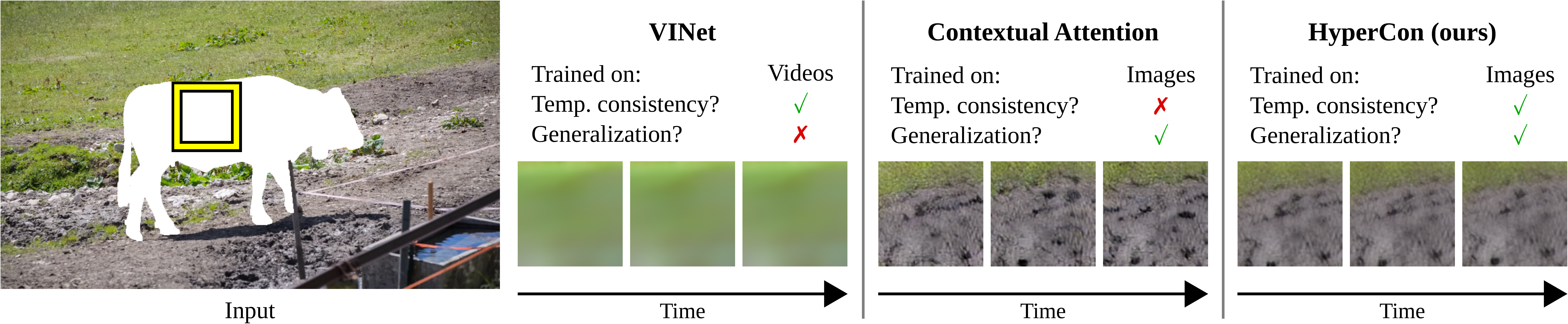}
  \caption{Video-to-video translation models designed and trained from scratch, \eg, VINet~\cite{Kim2019} are temporally consistent, but exhibit poor generalization performance due to the limited size of high-fidelity video datasets (note the lack of defined texture). Image-to-image models, \eg, Contextual Attention~\cite{Yu2018}, generalize well thanks to large image datasets, but lack temporal consistency (note the changing texture). HyperCon leverages the generalization performance conferred by image datasets while enforcing the temporal consistency properties of video-to-video models.}
  \label{fig:intro}
\end{figure*}

\vspace{-3pt}

\section{Introduction}
\label{sec:introduction}

Developments in both large-scale datasets and deep neural networks (DNNs) have led to incredible advancements in image-to-image~\cite{Isola2017,Yang2018} and video-to-video~\cite{Wang2018,Bansal2018,Park2019,Chen2019} translation tasks such as color restoration~\cite{Zhang2016,Zhang2017}, super-resolution~\cite{Dong2015a,Dong2015}, inpainting~\cite{Liu2018,Yu2018,Hong2019}, and style transfer~\cite{Gatys2016,Johnson2016}. But compared to images, videos pose an additional challenge: not only do the video frames need to satisfy the intended translation, they must also be temporally consistent. Otherwise, they will exhibit flickering artifacts.

Existing video-to-video translation techniques address temporal consistency through one of two types of strategies. The first incorporates temporal consistency directly into the method through optical flow-based losses~\cite{Kim2019,Ruder2016,Sajjadi2018,Gupta2017,Zhang2019} or network layers that operate on the time dimension, \eg, 3D convolutional layers~\cite{Chang2019} or recurrent layers~\cite{Kim2019}. These techniques leverage self-supervised video data and tailor models to specific tasks using relevant losses, \eg, reconstruction error~\cite{Liu2018,Yu2018,Kim2019} or style loss~\cite{Johnson2016,Gupta2017}. However, they require models and losses that are defined exclusively on videos and tuned for a specific application.

The other type of strategy uses a \emph{blind video consistency} model to reduce flicker in a frame-wise translated video as a post-processing step~\cite{Bonneel2015,Dong2015,Lai2018,Yao2017}. For example, Lai \etal~\cite{Lai2018} train a recurrent encoder-decoder network to improve the temporal consistency of independently processed frames via warping and structural preservation losses. Blind video consistency methods relax the need for task-specific video models and losses, and also enables image-to-image models to be applied immediately to videos without sacrificing consistency. However, they require dense correspondences between unprocessed frames during optimization, making them unsuitable for tasks in which certain input regions have no meaningful structure, \eg, video inpainting.

Similarly to blind video consistency methods, we opt to impart image-to-image models with temporal consistency, motivated primarily by generalization issues inherent to video-tailored approaches. To elaborate, consider in Figure~\ref{fig:intro} a failure case from the otherwise impressive state-of-the-art video inpainting network VINet~\cite{Kim2019}; here, it fails to hallucinate a sensible texture for the missing region. Now consider a state-of-the-art \textit{image} inpainting model, Contextual Attention~\cite{Yu2018}: this method produces realistic textures on the same example (but exhibits temporal inconsistency). The reason for this difference lies in the diversity of data used to train each model. Whereas the video model was trained on about 5,000 examples from one of the largest video segmentation datasets to date~\cite{Xu2018}, the image model was trained on over 1,000,000 images~\cite{Zhou2018}. Regardless of application, the vast scale of image datasets enables image models to encapsulate broader visual knowledge than video models trained from scratch and, as a result, better generalize to new data. Because storage and cost limitations make it intractable to collect high-quality video datasets as diverse as modern image datasets, video-tailored models are doomed to generalize poorly compared to image models.

To overcome this challenging generalization issue, we propose a method of \textit{image-to-video model transfer for video-to-video translation tasks}. Rather than optimize an image or video model from scratch, we aim to transform a black-box image-to-image translation model into a strong video-to-video translation model without fine-tuning---specifically, to automatically induce temporal consistency while achieving the same visual effect as the image-to-image model. Compared to prior techniques in blind video consistency~\cite{Bonneel2015,Lai2018}, which target these goals and therefore fall under the same problem space, ours broadens the scope of applicable tasks to include those in which the input and output videos have differing visual structure, such as video inpainting and edge-deforming style transfer (e.g., the \textit{mosaic} style from PyTorch's examples repository).

\begin{figure*}
  \centering
  \includegraphics[width=\linewidth]{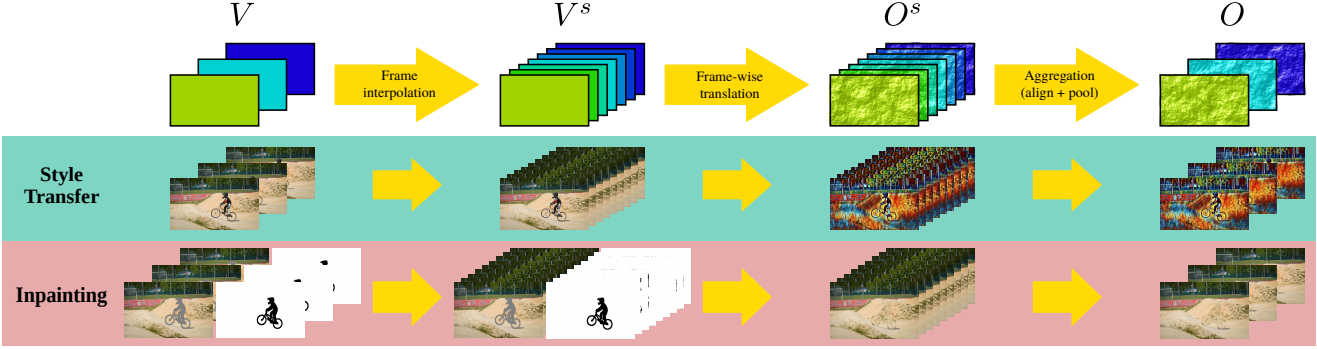}
  \caption{Visual overview of our Hyperconsistency (HyperCon) method. We begin by artificially inserting frames into the input video $V$ with a frame interpolation network to produce an interpolated video $V^s$. Then, we independently translate each frame in the interpolated video with an image-to-image translation model. Finally, we aggregate frames (\ie, align with optical flow and pool pixel-wise) within a local sliding window to produce the final temporally consistent output video $O$. This can be applied to tasks with or without masked inputs (\eg, inpainting and style transfer, respectively).}
  \label{fig:method-overview}
  \vspace{-3pt}
\end{figure*}

Key to our approach is the view of image-to-image models as noisy models in which small changes in the input lead to substantial changes in the output (\eg, Figure~\ref{fig:intro}, middle). To strengthen the desired signal and filter out noise, we synthesize several perturbed, but related predictions and aggregate over them. We obtain perturbations in a way that conditions on multiple neighboring frames to enhance temporal consistency. Our \emph{hyperconsistency} approach (\emph{HyperCon} for short) implements these principles by inserting frames into the video with a frame interpolation network, translating the interpolated video's frames independently, and aggregating within overlapping windows of appropriate stride to obtain a final video whose length matches the original (Figure~\ref{fig:method-overview}).

As the first method among image-to-video model transfer techniques to reason in interpolated video space, HyperCon forgoes the traditional post-processing paradigm by integrating frame-wise translation \emph{within} itself as an intermediate step, not as a step that precedes it. Since it does not require dense correspondences in the unprocessed input video, it performs well on video-to-video translation tasks with or without masked inputs---\eg, inpainting and style transfer respectively---as verified by our extensive experiments across these two widely differing applications. HyperCon outperforms a prior state-of-the-art video consistency model~\cite{Lai2018} in terms of reducing flicker and adhering to the intended translation; when combined with a strong image inpainting method, it also produces better predictions than a state-of-the-art video inpainting model~\cite{Kim2019}. It achieves competitive performance in both tasks \textit{despite not being trained with any masked or stylized videos}.

Our contributions are as follows. First, we motivate image-to-video model transfer as a way to leverage the superior generalization performance of image models for video-to-video translation without sacrificing temporal consistency. Second, we propose HyperCon, which supports a wider span of tasks than prior video consistency work thanks to its support for both masked \textit{and} unmasked inputs. Finally, we show that HyperCon performs favorably compared to state-of-the-art video consistency and inpainting methods without the need to be fine-tuned on these tasks. Our project website is available at \href{https://ryanszeto.com/projects/hypercon}{\texttt{ryanszeto.com/projects/hypercon}}.

\section{HyperCon}
\label{sec:tua}

Given an input video $V = \{v_1, \dots, v_N\}$, our goal is to generate an output video $O = \{o_1, \dots, o_N\}$ representing the $N$ frames of $V$ translated by some image-to-image model $g$. The frames of $O$ should closely resemble the frames of $V$ translated frame-wise by $g$; at the same time, $O$ should be temporally consistent, \ie, exhibit as few flickering effects as possible. We quantify these notions of resemblance and consistency concretely in Sections~\ref{sec:evaluation-metrics-style} and~\ref{sec:evaluation-metrics-inpainting}.

With HyperCon (Figure~\ref{fig:method-overview}), we generate $O$ as follows. First, we artificially insert $i$ frames between each pair of frames in $V$ with a frame interpolation network (Section~\ref{sec:generating-the-high-fps-video}). Denoting this interpolated version of $V$ as $V^s = \{v_1^s, \dots, v_{N'}^s\}$, where $N'$ is the number of frames in the interpolated video, we then independently translate frames in $V^s$ with $g$, yielding $O^s = \{o_1^s, \dots, o_{N'}^s\}$ (Section~\ref{sec:processing-the-high-fps-video}). Finally, we aggregate frames in $O^s$ over a temporal sliding window with appropriate stride to produce the frames of the final output video $O$ (Section~\ref{sec:temporal-aggregation}). Additional considerations for masked inputs are discussed in Section~\ref{sec:tua-for-masked-videos}.

\subsection{Generating the Interpolated Video}
\label{sec:generating-the-high-fps-video}

To generate the interpolated video $V^s$, we insert $i$ interpolated frames between each pair of frames in $V$, which essentially allows us to obtain several perturbed versions of each input frame for translation. We opt for a vector-based sampling method for frame interpolation instead of a kernel-based one\footnote{This distinction is discussed in more detail in Reda et al.~\cite{Reda2018}.} which, as we justify in Section~\ref{sec:tua-for-masked-videos}, allows us to handle the case of masked inputs appropriately. Specifically, for each pair of consecutive frames $v_a$ and $v_{a+1}$ ($a \in {1, \dots, N-1}$) and intermediate frame index $b \in \{1, \dots, i\}$, we predict two warping grids $(F^s_{a+b' \rightarrow a}, F^s_{a+b' \rightarrow a+1})$ and a weight map $w_{a+b'}$ (where $b' \equiv \frac{b+1}{i+1}$) with some function $\textrm{wrpgrd}$ (\eg, a pre-trained DNN), and use them to generate the corresponding interpolated frame $v_j^s$ ($j \in \{1, \dots, N'\}$):
\begin{gather}
  (F^s_{a+b' \rightarrow a}, F^s_{a+b' \rightarrow a+1}, w_{a+b'}) = \textrm{wrpgrd}(v_a, v_{a+1}, b') \, ,
  \label{eq:wrpgrd}
  \\
  v_j^s = (1 - w_{a+b'}) \odot \warp(v_a, F^s_{a+b' \rightarrow a})
  \nonumber
  \\
  \hspace{1.0cm} + w_{a+b'} \odot \warp(v_{a+1}, F^s_{a+b' \rightarrow a+1}) \, .
\end{gather}
$\odot$ is an element-wise product; $\textrm{warp}(v, F)$ bilinearly samples from $v$ via displacements specified by vector field $F$.

\subsection{Translating the Interpolated Video}
\label{sec:processing-the-high-fps-video}

At this point, we have computed the interpolated video $V^s$. We generate the translated interpolated video $O^s$ by simply translating each frame in $V^s$ independently:
\begin{align}
  o_j^s = g(v_j^s), \quad j \in \{1, \dots, N'\} \, .
\end{align}
Clearly, $O^s$ is not temporally consistent. However, we expect that most spatial regions in this video will exhibit consensus within small temporal windows. For example, a patch might have a distinct color profile in one frame, but a common color profile in the other frames in the local temporal window. Since we have more frames in $O^s$ than frames needed in the output, we can remove the spurious artifacts of frame-wise translation by mapping several neighboring frames in $O^s$ to one frame in our desired output video $O$. We call this mapping \textit{temporal aggregation} (Section~\ref{sec:temporal-aggregation}).

\subsection{Temporal Aggregation}
\label{sec:temporal-aggregation}

\begin{figure}
  \centering
  \includegraphics[width=\linewidth]{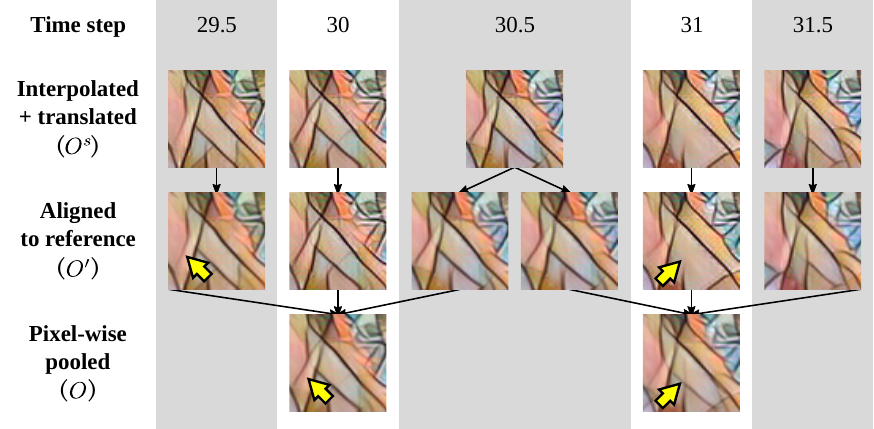}
  \caption{Temporal aggregation. Context frames from the translated interpolated video $O^s$ are aligned via optical flow to reference frames associated with integer time steps (white columns) and then pooled at each pixel location to generate the final video $O$. Despite inconsistencies between aligned frames (\eg, near the arrows), temporal aggregation selects stable components by majority vote.}
  \label{fig:temporal-aggregation}
\end{figure}

We perform temporal aggregation over a sliding window on the translated interpolated video $O^s$ (Figure~\ref{fig:temporal-aggregation}). The stride is such that the frame in each window's center, \ie, the reference frame, corresponds to a frame from the (non-interpolated) input video $V$, resulting in $N$ windows. Within each window, we align the off-center frames, \ie, the context frames, to the reference frame via optical flow warping, and then pool the reference and aligned context frames pixel-wise (\eg, with a mean or median filter) to produce a final frame in the output video $O$. Note that we compute the flow between \textit{interpolated, translated} frames $O^s$, \textit{not} the unprocessed input frames $V$ like prior work~\cite{Bonneel2015}.

More precisely, for an interpolated frame index $j \in \{1,$ $1+(i+1), \dots, N'-(i+1), N'\}$, we first estimate the optical flow between reference frame $o_j$ and each context frame in $\{ o_{j-d\gamma}^s, o_{j-d(\gamma-1)}^s, \dots,$ $o_{j-d}^s, o_{j+d}^s, \dots, o_{j+d(\gamma-1)}^s, o_{j+d\gamma}^s \}$ (denoted $F_{j \rightarrow (*)}^a$), where $\gamma$ and $d$ respectively parameterize the number of frames in the sliding window and a temporal dilation factor. We then warp the context frames to align them to $o_j^s$, and afterwards perform pixel-wise pooling over $o_j^s$ and the warped context frames:
\begin{align}
  o_{j,k}' &= \begin{cases}
    o_j^s & k = 0
    \\
    \warp(o_{j+dk}^s, F^a_{j \rightarrow j+dk}) & k \neq 0
  \end{cases} , k \in \{ -\gamma, \dots, \gamma \} \, ,
  \\
  o_j &= \textrm{pool} \big( o_{j, k}' \mid k \in \{ -\gamma, \dots, \gamma \} \big) \, .
\end{align}
Pooling is applied over values per spatial location, color channel, and time step; \ie, if $P = \textrm{pool}(I_1, I_2, \dots)$, then
\begin{align}
  P(l_h, l_w, l_c) = f\big( I_1(l_h, l_w, l_c), I_2(l_h, l_w, l_c), \dots \big) \, ,
\end{align}
where $P(l_h, l_w, l_c)$ and $I(l_h, l_w, l_c)$ denote the value of 3D image tensors $P$ and $I$ at location $(l_h, l_w, l_c)$, and $f$ is a mean or median operation. In the cases where the sliding window samples outside the valid frame range, we only align and pool over valid frames.

To illustrate why HyperCon induces temporal consistency, we visualize intermediate outputs for style transfer in Figure~\ref{fig:temporal-aggregation}. Even among interpolated frames with similar appearances, flickering artifacts can occur. By selecting pixel values by a majority vote over several interpolated frames, our method automatically incorporates stable components into the final prediction, thereby reducing flicker.

\subsection{HyperCon for Masked Videos}
\label{sec:tua-for-masked-videos}

Having described HyperCon in the unmasked input case in Sections~\ref{sec:generating-the-high-fps-video}-\ref{sec:temporal-aggregation}, we now extend it to handle tasks in which the input frames have masked pixels (\eg, inpainting). This case differs from the unmasked input case in three ways. First, in addition to the normal RGB video $V$, we now have as input a mask video $M = \{m_1, \dots, m_N\}$ in which 1 marks an unmasked pixel and 0 marks a masked pixel. Second, when generating the interpolated data, we must create an interpolated mask video $M^s = \{m_1^s, \dots, m_{N'}^s\}$ to accompany the interpolated RGB video $V^s$. Finally, the image-to-image translation model $g$ now takes a mask as input in addition to an RGB video frame.

We modify the interpolated video generation step (Section~\ref{sec:generating-the-high-fps-video}) to produce both $V^s$ and $M^s$; this is done by generating $i$ interpolated frames between each pair of frames in $V$ and $M$. For this to be valid, the motion of the interpolated mask video must match that of the interpolated RGB video---for example, if we interpolate the motion of a removed person, the mask must cover that person throughout the interpolated sequence. If this is not handled properly, we risk polluting the final result with mask placeholder values. Thus, we opt for a vector-based sampling method for frame interpolation instead of a kernel-based one, since the same warping grid can be applied to both RGB and mask frames to achieve the desired result.

To generate $V^s$, recall that we predict warping grids and a weight map $(F_{a+b' \rightarrow a}^s, F_{a+b' \rightarrow a+1}^s, w_{a+b'})$ from frames in $V$ using Equation~\ref{eq:wrpgrd}. To obtain the interpolated masks $M^s$, we apply these parameters to the masks in $M$ and follow up with a thresholding operation:
\begin{align}
  \dot{m}_j^s &= (1 - w_{a+b'}) \odot \warp(m_a, F_{a+b' \rightarrow a}^s)
  \nonumber
  \\
  &\quad + w_{a+b'} \odot \warp(m_{a+1}, F_{a+b' \rightarrow a+1}^s) \, ,
  \\
  m_j^s &= \thresh(\dot{m}_j^s, 1) \, .
\end{align}
Warping the masks in this way allows us to detect the ``partially-masked'' pixels in $v_j^s$, \ie, the ones that received a contribution from a masked pixel in either $v_a$ or $v_{a+1}$. Specifically, if a pixel in $\dot{m}_j^s$ is not 1, then the warping operation used a source value of 0 from $m_a$ or $m_{a+1}$, which corresponds to borrowing from a masked pixel. Thus, thresholding turns partially-masked pixels into fully-masked pixels in the interpolated masks so that the subsequent translation step is not incorrectly influenced by these pixels.

At this point, we have generated the interpolated RGB and mask videos $V^s$ and $M^s$. We apply the image-to-image model $g$ to them:
\begin{align}
  o_j^s = g(v_j^s, m_j^s), \quad j \in {1, \dots, N'} \, ,
\end{align}
and then apply temporal aggregation (Section~\ref{sec:temporal-aggregation}) as usual.

\subsection{HyperCon Implementation Details}
\label{sec:hypercon-implementation-details}

For the frame interpolation step, we use Super SloMo~\cite{Jiang2018} as our wrpgrd function to predict warping grids and weight masks. This method is well-suited for our approach since it predicts warping parameters for multiple intermediate time steps, contrasting with kernel-based sampling methods that interpolate one frame~\cite{Niklaus2017}. To estimate optical flow in the temporal aggregation step, we use a third-party implementation of PWC-Net~\cite{Sun2017}. Since HyperCon is agnostic to the specific instantiations of frame interpolation and flow estimation, we have chosen state-of-the-art models to demonstrate the full potential of our overall approach. Although we do not fine-tune network weights, we observe good performance regardless and postulate that performance can be further improved by fine-tuning.

\section{Experiments: Video Style Transfer}
\label{sec:experiments-video-style-transfer}

To demonstrate HyperCon on unmasked inputs, we apply it to video style transfer. For the frame-wise style transfer subroutine, we use the Fast Style Transfer (FST) models from Johnson \etal~\cite{Johnson2016} with the pre-trained \textit{mosaic} and \textit{rain-princess} weights from PyTorch's examples repository.

\subsection{Datasets}
\label{sec:datasets-style}

For evaluation, we use the ActivityNet~\cite{Heilbron2015} and DAVIS~\cite{Perazzi2016} video datasets, which primarily consist of dynamic indoor and outdoor scenes of animals and people. We split them into validation and test sets to validate HyperCon hyperparameters and evaluate performance, respectively. For ActivityNet, we manually curate clips from the official training/validation and test splits ({\raise.17ex\hbox{$\scriptstyle\sim$}}100 videos per split with {\raise.17ex\hbox{$\scriptstyle\sim$}}100 frames per video), filtering by properties of high-quality videos such as non-negligible motion, no splash screens, one continuous camera shot, \etc. For DAVIS, we use the official training/validation set for validation, and combine the development and challenge sets from the DAVIS 2017 and 2019 challenges to produce two test sets, one for each year. We pre-process all videos by resizing and center-cropping them to 832$\times$480 resolution, and scaling RGB values to \mbox{(-1, 1)}.

\subsection{Evaluation Metrics}
\label{sec:evaluation-metrics-style}

For video style transfer, our goal is to transfer the desired style to all video frames while minimizing flickering effects (\ie, maximizing temporal consistency) in the output video as much as possible.
To evaluate temporal consistency, we measure warping error $E_{\textrm{warp}}$~\cite{Lai2018} and patch-based consistency measures $C_{\textrm{PSNR}}$ and $C_{\textrm{SSIM}}$~\cite{Gupta2017}. $E_{\textrm{warp}}$ is defined as the mean of $e_{\textrm{warp}}$ over all consecutive pairs of output frames $(o_a, o_{a+1})$, where
\begin{align}
  e_{\textrm{warp}}(o_a, o_{a+1}) = \frac{1}{\sum_p M_a^f(p)} \sum_p M_a^f(p) \| D_a(p) \|_2^2 \, .
\end{align}
Here, $D_a = o_a - \textrm{warp}(o_{a+1}, F_{a \rightarrow a+1})$ (where $F_{a \rightarrow a+1}$ is the estimated flow between input frames $v_a$ and $v_{a+1}$); $p$ indexes the pixels in the frame; and $M_a^f$ is a mask that indicates pixels with reliable flow (1 for reliable, 0 for unreliable). $M_a^f$ is computed based on flow consistency and motion boundaries as defined by Ruder \etal~\cite{Ruder2016}. $C_{\textrm{PSNR}}$ is defined as the mean of $c_{\textrm{PSNR}}$ over all consecutive pairs of output frames, where $c_{\textrm{PSNR}}$ is computed by taking a random 50$\times$50 patch in frame $o_a$ and computing the maximum PSNR between it and all patches in its spatial neighborhood in the next frame $o_{a+1}$ (within a Chebyshev distance of 20 pixels). $C_{\textrm{SSIM}}$ is defined similarly to $C_{\textrm{PSNR}}$, except it computes Structural Similarity~\cite{Wang2004} in place of PSNR. To quantify how well a method adheres to the intended style, we measure Frech\'et Inception Distance (FID)~\cite{Heusel2017} between the set of all frames generated by frame-wise translation and the set of all frames generated by the method under evaluation.

\begin{table*}
  \centering
  \scalebox{0.7}{
    \begin{tabular}{c|c|cccc|cccc}
    \hline
     &  & \multicolumn{4}{c|}{mosaic} & \multicolumn{4}{c}{rain-princess} \\
    \hline
    Dataset & Method & FID$^\downarrow$ & $E_{\textrm{warp}}$$^\downarrow$ & $C_{\textrm{PSNR}}$$^\uparrow$ & $C_{\textrm{SSIM}}$$^\uparrow$ & FID$^\downarrow$ & $E_{\textrm{warp}}$$^\downarrow$ & $C_{\textrm{PSNR}}$$^\uparrow$ & $C_{\textrm{SSIM}}$$^\uparrow$ \\
    \hline
    \multirow{3}{*}{DAVIS 2017} & FST~\cite{Johnson2016} & - & 0.024829 ± 0.001174 & 16.72 ± 1.43 & 0.5166 ± 0.0152 & - & 0.013934 ± 0.000903 & 19.75 ± 1.39 & 0.6030 ± 0.0156 \\
     & FST-vcons~\cite{Lai2018} & 24.50 & 0.010194 ± 0.000500 & 20.51 ± 1.37 & 0.5830 ± 0.0140 & 10.87 & 0.007071 ± 0.000477 & 22.73 ± 1.35 & 0.6717 ± 0.0136 \\
     & HyperCon (ours) & \textbf{18.04} & \textbf{0.008810 ± 0.000493} & \textbf{21.04 ± 1.37} & \textbf{0.6662 ± 0.0157} & \textbf{10.53} & \textbf{0.006110 ± 0.000487} & \textbf{23.38 ± 1.35} & \textbf{0.7480 ± 0.0143} \\
    \hline
    \multirow{3}{*}{DAVIS 2019} & FST~\cite{Johnson2016} & - & 0.029379 ± 0.001684 & 14.88 ± 0.29 & 0.4737 ± 0.0184 & - & 0.017614 ± 0.001309 & 17.61 ± 0.38 & 0.5550 ± 0.0189 \\
     & FST-vcons~\cite{Lai2018} & 24.89 & 0.011999 ± 0.000672 & 18.79 ± 0.29 & 0.5451 ± 0.0169 & 14.77 & 0.008938 ± 0.000644 & 20.86 ± 0.38 & 0.6365 ± 0.0159 \\
     & HyperCon (ours) & \textbf{23.11} & \textbf{0.010677 ± 0.000710} & \textbf{19.20 ± 0.32} & \textbf{0.6105 ± 0.0192} & \textbf{13.59} & \textbf{0.008117 ± 0.000730} & \textbf{21.13 ± 0.43} & \textbf{0.6960 ± 0.0175} \\
    \hline
    \multirow{3}{*}{ActivityNet} & FST~\cite{Johnson2016} & - & 0.017895 ± 0.000847 & 19.29 ± 0.81 & 0.6478 ± 0.0134 & - & 0.008428 ± 0.000516 & 23.45 ± 0.81 & 0.7469 ± 0.0116 \\
     & FST-vcons~\cite{Lai2018} & 26.37 & 0.006727 ± 0.000311 & 22.58 ± 0.38 & 0.7075 ± 0.0115 & 9.07 & 0.003923 ± 0.000240 & 25.97 ± 0.42 & 0.8008 ± 0.0093 \\
     & HyperCon (ours) & \textbf{10.09} & \textbf{0.005946 ± 0.000298} & \textbf{23.61 ± 0.77} & \textbf{0.7848 ± 0.0102} & \textbf{5.50} & \textbf{0.003379 ± 0.000233} & \textbf{26.95 ± 0.76} & \textbf{0.8544 ± 0.0083} \\
    \hline
    \end{tabular}
  }
  \caption{Quantitative comparison between frame-wise style transfer (FST)~\cite{Johnson2016}, baseline blind video consistency (FST-vcons)~\cite{Lai2018}, and HyperCon (ours) for style transfer. $^\uparrow$ and $^\downarrow$ indicate where higher and lower is better; bold indicates the best score; and the values after $\pm$ are standard error over all videos. FID for FST is blank since this equates to comparing FST to itself. HyperCon obtains lower FID scores than FST-vcons, indicating greater coherence to the intended style of FST, as well as better warping errors and patch-based consistency scores, indicating better temporal consistency.
  }
  \label{tab:baselines-style}
\end{table*}

\subsection{Hyperparameter Analysis}
\label{sec:parameter-setting-analysis}

\begin{figure}
  \centering
  \includegraphics[width=\linewidth]{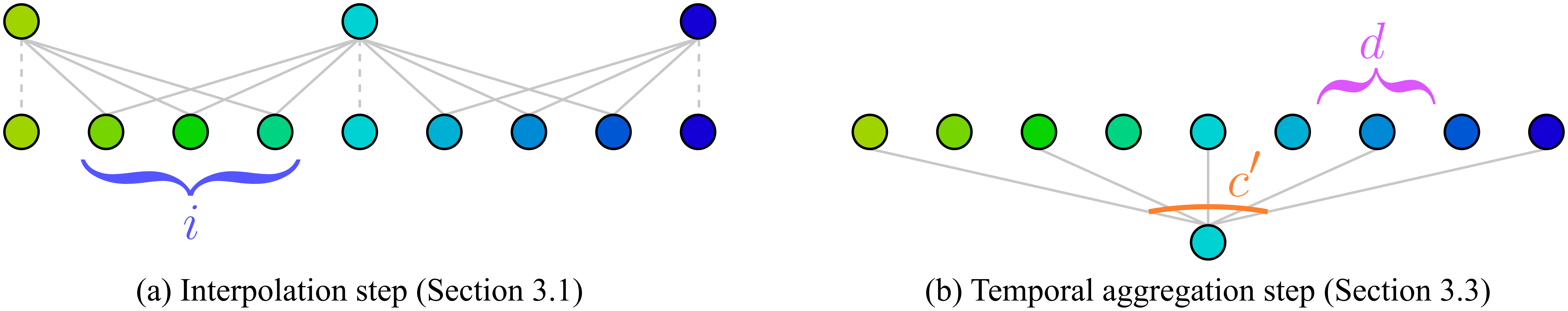}
  \caption{Visualization of the hyperparameters included in our grid search (Section~\ref{sec:parameter-setting-analysis}).}
  \label{fig:grid-search}
\end{figure}

\begin{figure}
  \centering
  \begin{subfigure}{\linewidth}
    \centering
    \includegraphics[width=0.95\linewidth]{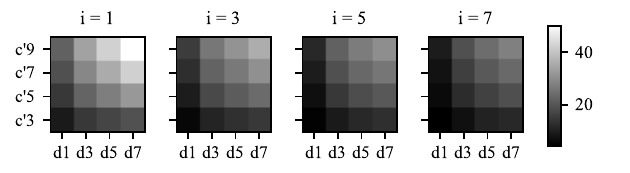}
  \end{subfigure}
  \begin{subfigure}{\linewidth}
    \centering
    \includegraphics[width=0.95\linewidth]{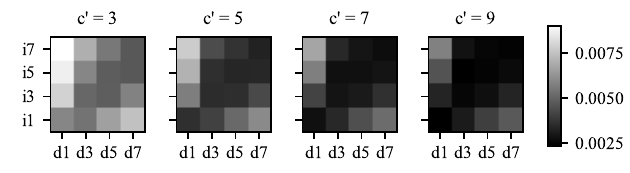}
  \end{subfigure}
  \caption{Hyperparameter grid search for \textit{rain-princess} on the DAVIS train/val set. Row 1: FID (style adherence). Row 2: $E_{\textrm{warp}}$ (temporal consistency). Lower is better.}
  \label{fig:parameter-setting-quantitative}
  \vspace{-8pt}
\end{figure}

We perform a grid search over our method's hyperparameters to link them concretely to style adherence and temporal consistency. These hyperparameters consist of the number of frames to insert between each pair $i$, the total number of interpolated frames to aggregate for each output frame $c' = 2\gamma + 1$, and the spacing between aggregated interpolated frames $d$ (Figure~\ref{fig:grid-search}). We quantify each setting's performance and present their trends in Figure~\ref{fig:parameter-setting-quantitative}.

\begin{figure*}
  \centering
  \begin{subfigure}{0.49\linewidth}
    \begin{subfigure}{0.32\linewidth}
      \includegraphics[width=\linewidth]{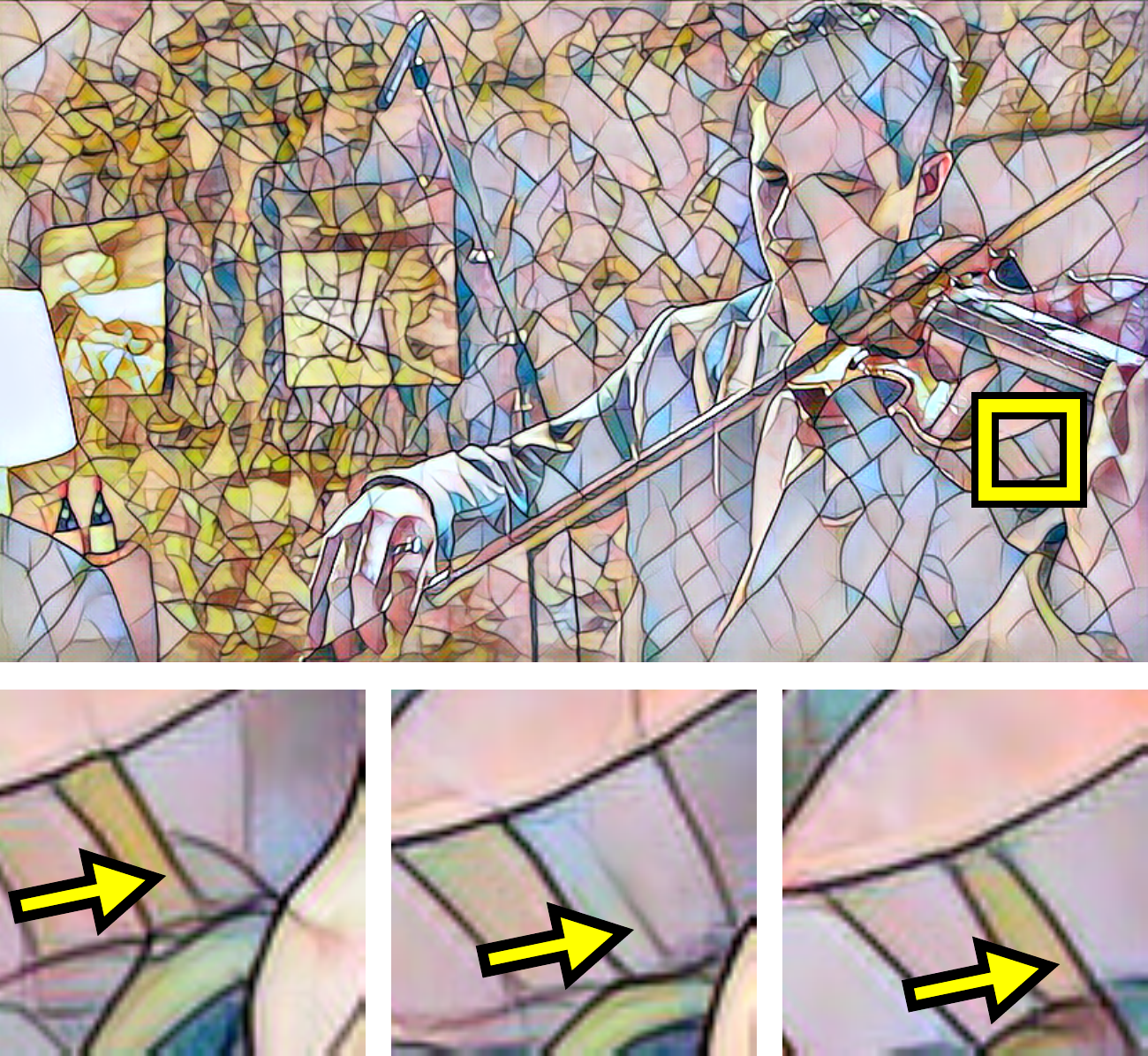}
      \caption{FST~\cite{Johnson2016}}
      \label{fig:baselines-style-mosaic-fst}
    \end{subfigure}
    \hfill
    \begin{subfigure}{0.32\linewidth}
      \includegraphics[width=\linewidth]{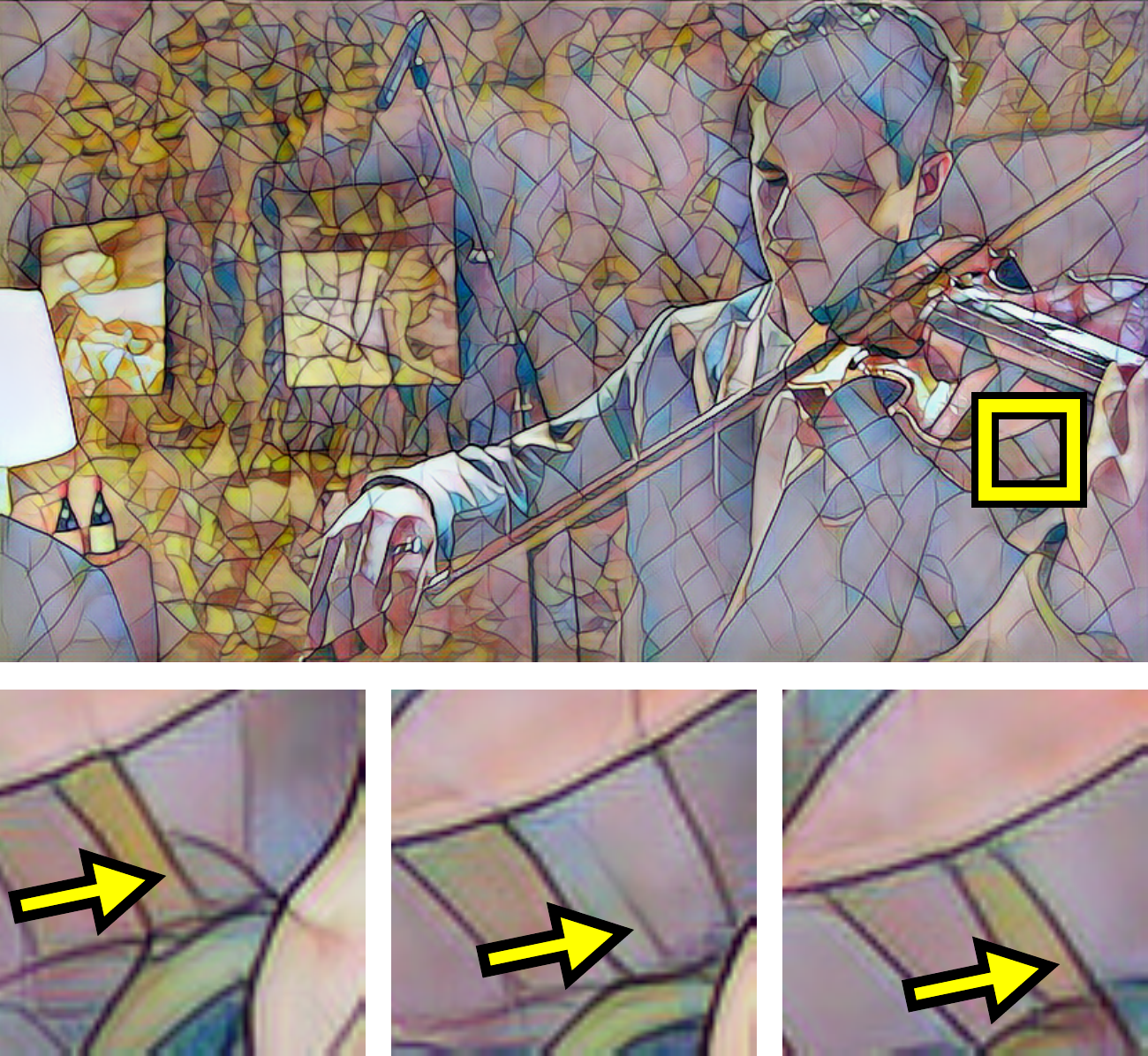}
      \caption{FST-vcons~\cite{Lai2018}}
      \label{fig:baselines-style-mosaic-fst-vcons}
    \end{subfigure}
    \hfill
    \begin{subfigure}{0.32\linewidth}
      \includegraphics[width=\linewidth]{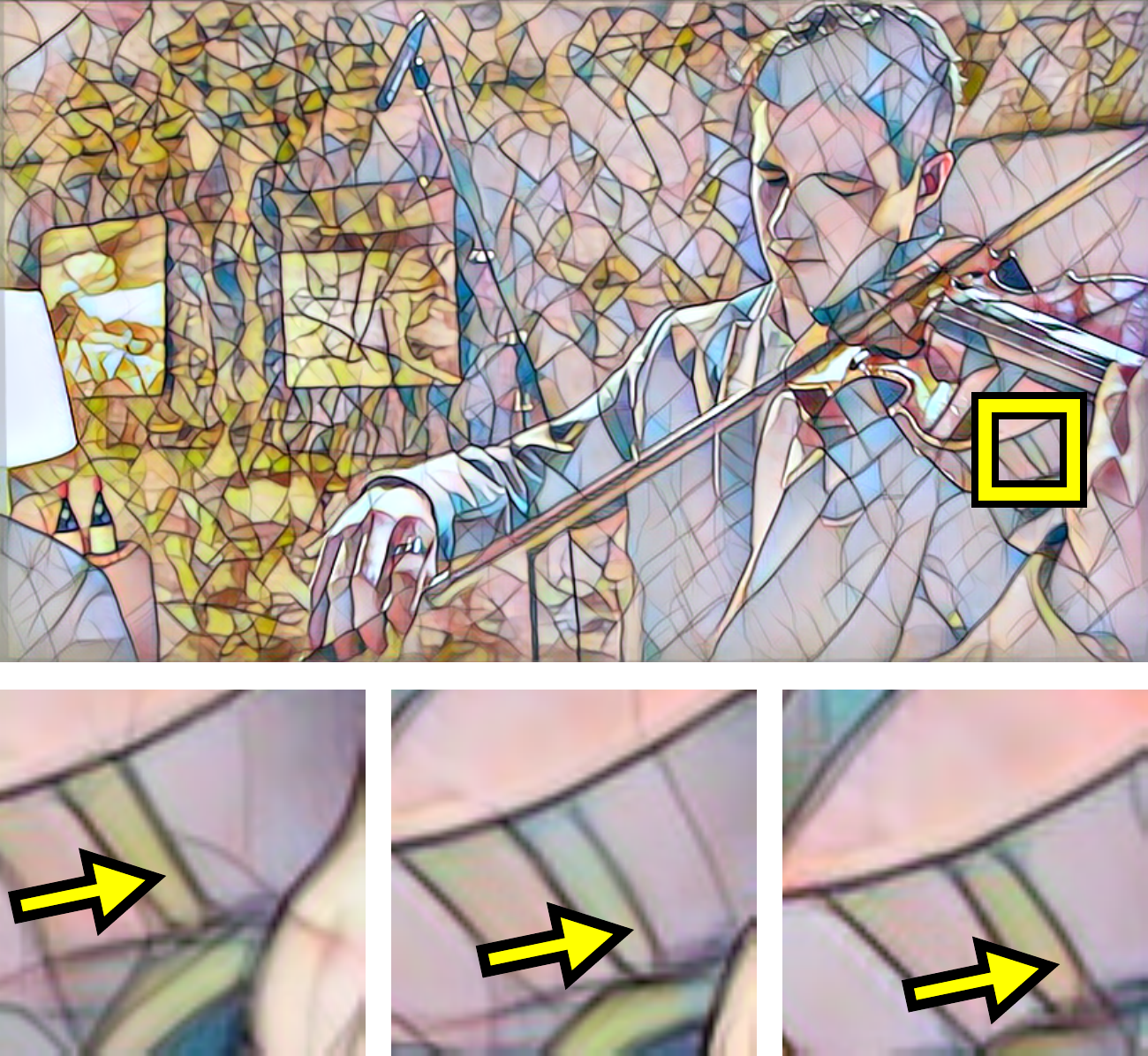}
      \caption{HyperCon (ours)}
      \label{fig:baselines-style-mosaic-tua}
    \end{subfigure}
  \end{subfigure}
  \hfill
  \begin{subfigure}{0.49\linewidth}
    \begin{subfigure}{0.32\linewidth}
      \includegraphics[width=\linewidth]{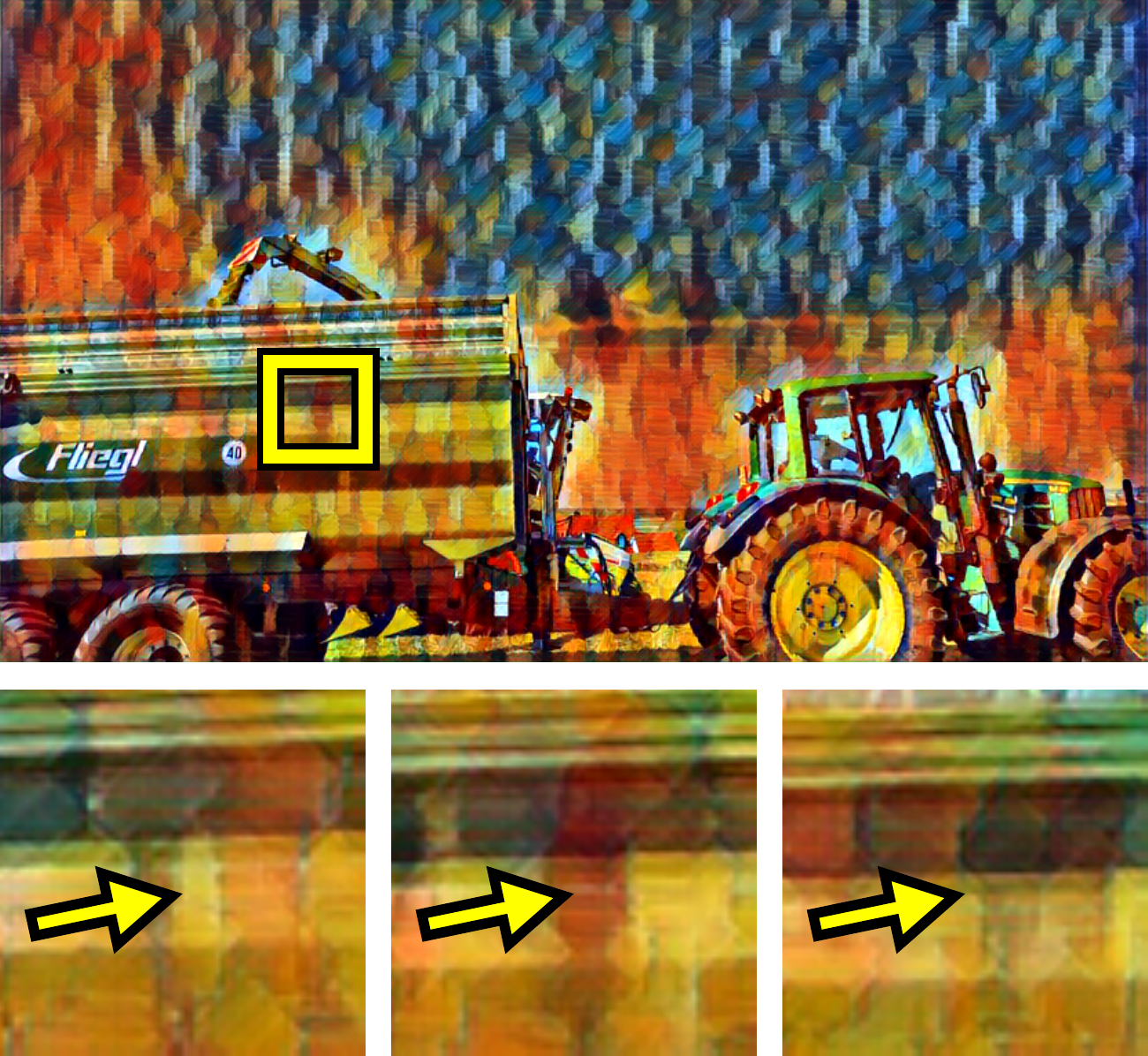}
      \caption{FST~\cite{Johnson2016}}
      \label{fig:baselines-style-rain-princess-fst}
    \end{subfigure}
    \hfill
    \begin{subfigure}{0.32\linewidth}
      \includegraphics[width=\linewidth]{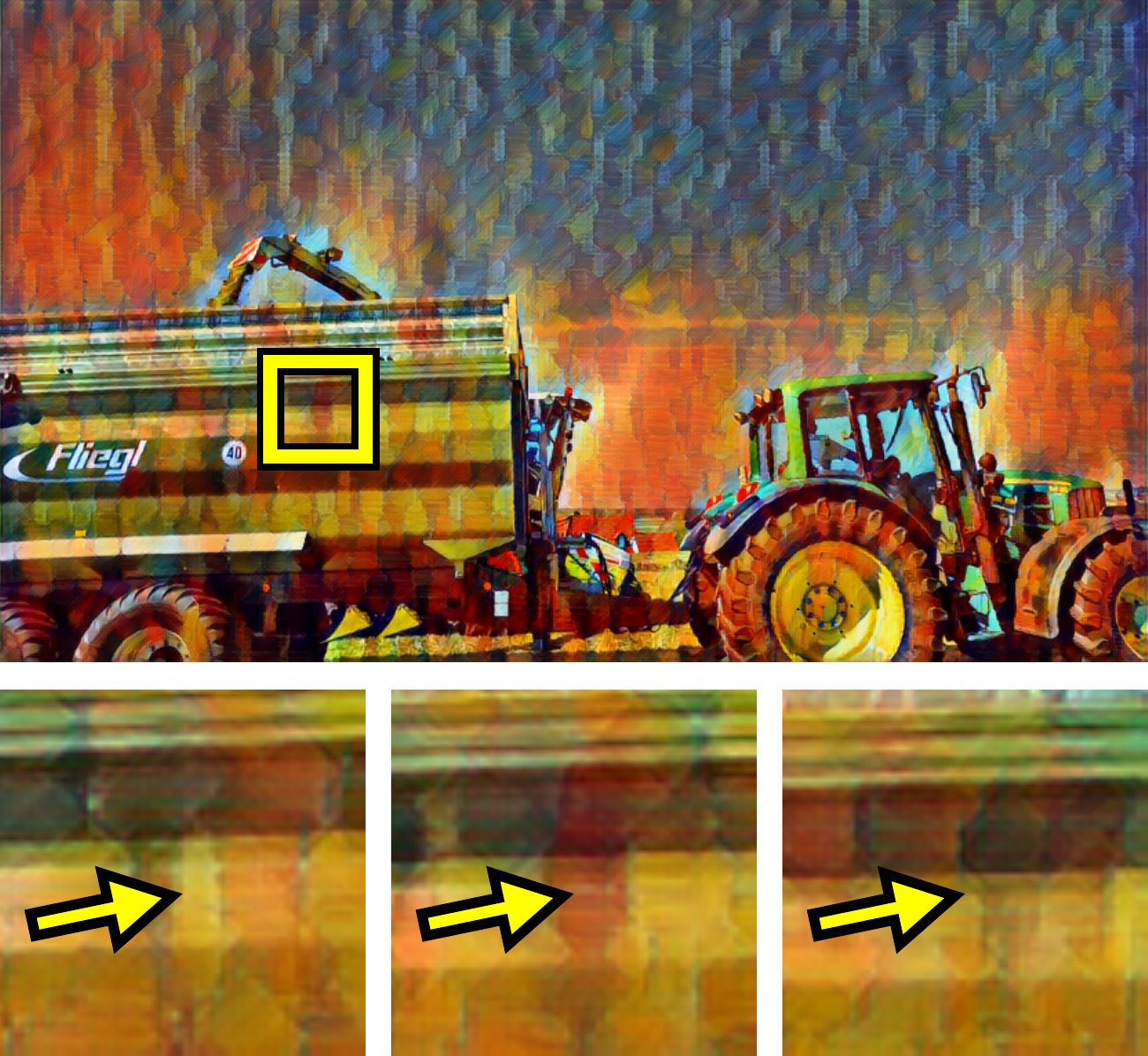}
      \caption{FST-vcons~\cite{Lai2018}}
      \label{fig:baselines-style-rain-princess-fst-vcons}
    \end{subfigure}
    \hfill
    \begin{subfigure}{0.32\linewidth}
      \includegraphics[width=\linewidth]{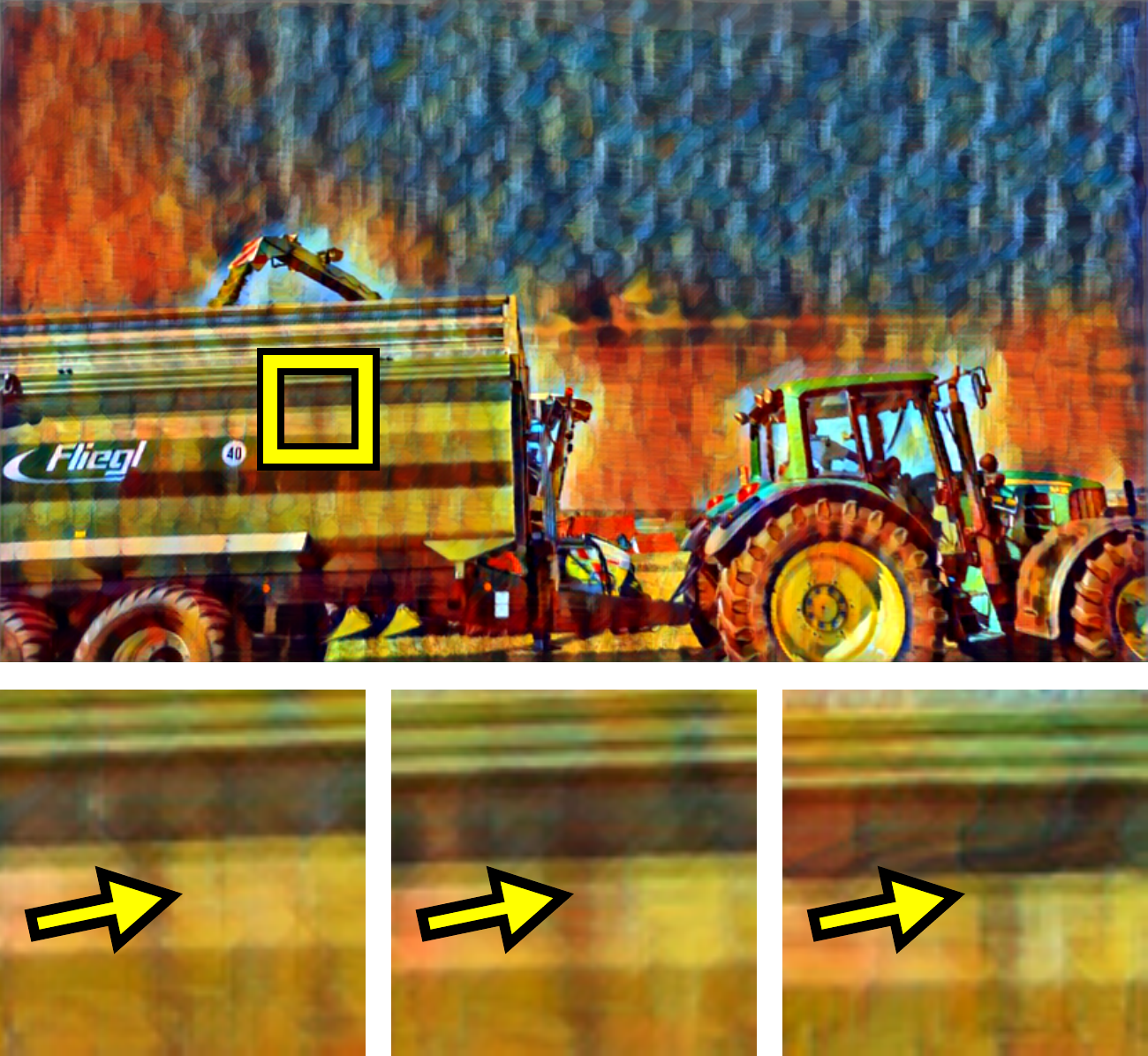}
      \caption{HyperCon (ours)}
      \label{fig:baselines-style-rain-princess-tua}
    \end{subfigure}
  \end{subfigure}
  \caption{Style transfer comparison for \textit{mosaic} (left) and \textit{rain-princess} (right) styles. We show one full frame and crops from three consecutive frames centered at the presented frame. Unlike the baselines, HyperCon draws three consistent lines across the top of the violin (left), and removes the flickering spot on the truck (right), thus producing temporally consistent results.}
  \label{fig:baselines-style}
  \vspace{-3pt}
\end{figure*}

Lai et al.~\cite{Lai2018} observe that style adherence and temporal consistency are competing objectives; our hyperparameter grid search supports this claim. Specifically, FID decreases with more interpolated frames, fewer aggregated frames, and a smaller dilation rate, indicating that style adherence is maximized when HyperCon aggregates over few frames that are very similar to the reference frame. On the other hand, $E_{\textrm{warp}}$ decreases gradually with more aggregated frames, and follows the trend of a parabolic cylinder given fixed $c'$, \ie, there is a sweet spot for $i$ given fixed $d$ and vice-versa. The trends of $E_{\textrm{warp}}$ suggest that temporal consistency is maximized when many frames are aggregated over some effective frame rate. Due to the inherent trade-off between style adherence and temporal consistency, our hyperparameter selection process considers the strongest models in the former criterion and chooses the one that best satisfies the latter; specifically, among the models ranked in the top 15\% by FID, we select the one that minimizes $E_{\textrm{warp}}$.

\subsection{Comparison To Prior State-of-the-Art}
\label{sec:comparison-to-prior-sota-style}

Now we compare HyperCon to frame-wise style transfer (FST)~\cite{Johnson2016} and a state-of-the-art video consistency method from Lai \etal{} (FST-vcons)~\cite{Lai2018}. For FST-vcons, we first apply FST to the video, and then apply the consistency model of Lai \etal{} as a post-translation step. Note that FST-vcons must process inputs sequentially, whereas HyperCon can operate on several time steps in parallel due to its short-range dependencies. Given efficient implementations of both methods, HyperCon's parallelizability gives it an advantage in terms of inference time on long videos.

We provide a quantitative comparison of these methods in Table~\ref{tab:baselines-style}. Since FST stylizes each frame independently, it naturally yields the worst temporal consistency as indicated by its relatively poor warping errors and patch-based consistency scores. Between FST-vcons and our HyperCon approach, the latter obtains better FID scores, warping errors, and patch-based consistency scores, indicating that it is more temporally consistent and better captures the intended style and content compared to FST-vcons.

To ensure that these conclusions match those derived from human judgements, we devise a survey on Amazon Mechanical Turk to compare HyperCon and the FST-vcons baseline on our test videos (Section~\ref{sec:datasets-style}). To judge overall quality, we present subjects with corresponding videos from each method and ask which is more appealing---the percentage of responses that favor our method is \textbf{57.3\%} for the rain-princess style and \textbf{55.5\%} for the mosaic style. As for style adherence, we present the two aforementioned videos alongside the frame-wise translated one from FST, and ask method is more similar to FST---here, the percentage of responses that favor our method is \textbf{79.1\%} for rain-princess and \textbf{76.8\%} for mosaic. Survey design details are included in the supplementary materials.

\begin{table*}
  \centering
  \scalebox{0.64}{
    \begin{tabular}{c|ccccc|ccccc|ccccc}
      \hline
       & \multicolumn{5}{c|}{DAVIS 2017} & \multicolumn{5}{c|}{DAVIS 2019} & \multicolumn{5}{c}{ActivityNet} \\
      \hline
      Method & $D_{\textrm{LPIPS}}$$^\downarrow$ & $D_{\textrm{VLPIPS}}$$^\downarrow$ & FID$^\downarrow$ & VFID$^\downarrow$ & $E_{\textrm{warp}}$$^\downarrow$ & $D_{\textrm{LPIPS}}$$^\downarrow$ & $D_{\textrm{VLPIPS}}$$^\downarrow$ & FID$^\downarrow$ & VFID$^\downarrow$ & $E_{\textrm{warp}}$$^\downarrow$ & $D_{\textrm{LPIPS}}$$^\downarrow$ & $D_{\textrm{VLPIPS}}$$^\downarrow$ & FID$^\downarrow$ & VFID$^\downarrow$ & $E_{\textrm{warp}}$$^\downarrow$ \\
      \hline
      Cxtattn~\cite{Yu2018} & 0.0457 & 0.5838 & 20.94 & 1.435 & 0.002186 & 0.0442 & 0.5575 & \textbf{15.55} & 1.361 & 0.002539 & \textubf{0.0432} & 0.5981 & \textubf{21.5173} & 1.4417 & 0.000894 \\
      Cxtattn-vcons~\cite{Lai2018} & 0.0480 & 0.6076 & 23.23 & 1.502 & 0.001780 & 0.0478 & 0.5964 & 18.52 & 1.490 & 0.002166 & 0.0448 & 0.6067 & 23.1642 & 1.4383 & 0.000689 \\
      VINet~\cite{Kim2019} & 0.0616 & 0.6062 & 29.24 & 1.465 & 0.001882 & 0.0539 & 0.5455 & 18.22 & \textubf{1.195} & 0.002292 & 0.0608 & 0.6139 & 29.3806 & 1.3783 & \textbf{0.000678} \\
      HyperCon-mean (ours) & \textbf{0.0450} & \textbf{0.5272} & \textbf{18.49} & \textubf{1.073} & \textubf{0.001540} & \textbf{0.0437} & \textbf{0.5179} & 16.07 & 1.274 & \textubf{0.001847} & 0.0454 & \textubf{0.5728} & 22.5111 & \textubf{1.2251} & \textubf{0.000640} \\
      HyperCon-median (ours) & \textubf{0.0424} & \textubf{0.5217} & \textubf{17.75} & \textbf{1.074} & \textbf{0.001614} & \textubf{0.0419} & \textubf{0.5089} & \textubf{15.24} & \textbf{1.254} & \textbf{0.001950} & \textbf{0.0441} & \textbf{0.5812} & \textbf{22.2705} & \textbf{1.2601} & 0.000683 \\
      \hline
    \end{tabular}
  }
  \caption{Comparison between baseline methods and HyperCon (ours) on the simulated video inpainting task. $^\downarrow$ indicates that lower is better. Bold+underline and bold respectively indicate the 1st- and 2nd-place methods. Among the evaluated methods, the HyperCon models consistently place in the top two across all performance measures.}
  \label{tab:swr-sor-quantitative}
  \vspace{-6pt}
\end{table*}

In Figure~\ref{fig:baselines-style}, we visually compare FST, FST-vcons, and HyperCon on two DAVIS 2017 test videos. Naturally, FST generates temporally inconsistent predictions by operating on each frame independently: observe the changing arrangement of lines across the top of the violin in Figure~\ref{fig:baselines-style-mosaic-fst}, as well as the flashing red spot in Figure~\ref{fig:baselines-style-rain-princess-fst}. Meanwhile, FST-vcons has two systematic failures. First, it greatly desaturates predictions, observable across the full frame in Figure~\ref{fig:baselines-style-mosaic-fst-vcons} and in the dulled orange and blue hues of the sky in Figure~\ref{fig:baselines-style-rain-princess-fst-vcons}. Second, it leaves inconsistencies intact as a result of darkening regions instead of shifting their hue; for instance, in Figure~\ref{fig:baselines-style-mosaic-fst-vcons}, the pattern of lines on the violin match the inconsistent appearance of FST. HyperCon, on the other hand, properly addresses both of these challenges---in Figure~\ref{fig:baselines-style-mosaic-tua}, HyperCon maintains a consistent pattern of violin lines by passing relevant information between frames, and in Figure~\ref{fig:baselines-style-rain-princess-tua}, it removes the flashing red spot without desaturating the rest of the frame like FST-vcons.

\section{Experiments: Video Inpainting}
\label{sec:experiments-video-inpainting}

To evaluate HyperCon on masked inputs, we apply it to video inpainting. For the frame inpainting subroutine, we re-train the Contextual Attention model from Yu \etal~\cite{Yu2018} using a modified training scheme that yields higher-quality predictions. Specifically, whereas Yu \etal{} use the WGAN-GP formulation~\cite{Gulrajani2017} to update the discriminators of their adversarial loss, we use the original cross-entropy GAN formulation~\cite{Goodfellow2014} with spectral normalization layers~\cite{Miyato2018} in the discriminator, which stabilize GAN training. Our model achieves a PSNR of 20.41 dB on the Places dataset~\cite{Zhou2018}, surpassing the original reported performance of 18.91 dB~\cite{Yu2018} under the same evaluation setting.

\subsection{Datasets}
\label{sec:datasets-inpainting}

To evaluate video inpainting methods quantitatively, we automatically synthesize occlusion masks and use the method under evaluation to inpaint the masked area. As in prior work~\cite{Wang2019,Xu2019}, our masks take the form of a rectangle at a fixed location throughout time. For each video, we randomly select the corner locations of the rectangle, restricting its height and width to lie between 15-50\% of the full frame's dimensions (the masks are the same for all evaluated methods). We use the same videos for hyperparameter tuning and evaluation as those described in Section~\ref{sec:datasets-style}.

\subsection{Evaluation Metrics}
\label{sec:evaluation-metrics-inpainting}

The performance measures that we use for inpainting broadly assess temporal consistency, reconstruction quality, and realism of the composited frames, \ie, the predicted frames with unoccluded pixels replaced by those from the input. We measure temporal consistency using $E_{\textrm{warp}}$ as described in Section~\ref{sec:evaluation-metrics-style}. For reconstruction quality, we report the mean LPIPS distance ($D_{\textrm{LPIPS}}$)~\cite{Zhang2018} between corresponding composited and ground-truth (\ie, unoccluded) frames. Although traditional quality measures such as PSNR and SSIM~\cite{Wang2004} have been used in prior image inpainting work~\cite{Yu2018,Liu2018}, Zhang \etal~\cite{Zhang2018} have shown that LPIPS better correlates with human judgment for paired image comparisons. Additionally, to capture spatio-temporal similarity, we define a new video-based version of LPIPS, VLPIPS, which is a distance between several layers of activations from the I3D action recognition network~\cite{Carreira2017}. We exhaustively compute VLPIPS between corresponding 10-frame segments from the composited and ground-truth videos and report the overall mean. Finally, to evaluate realism, we report FID~\cite{Heusel2017} between the sets of composited and ground-truth frames, as well as a video variant VFID~\cite{Wang2018} between the sets of 10-frame segments from the composited and ground-truth videos. We compute VFID using the output of I3D's final average pooling layer.

\subsection{Comparison to Prior State-of-the-Art}
\label{sec:comparison-to-prior-sota-inpainting}

For inpainting, we consider two variants of our model: one using a mean filter during temporal aggregation (Section~\ref{sec:temporal-aggregation}), and the other using a median filter. We compare our models to three state-of-the-art baselines. The first method, Contextual Attention (Cxtattn), inpaints frames independently using the model from Yu \etal~\cite{Yu2018} with our improved weights (Section~\ref{sec:experiments-video-inpainting}). The second method, Cxtattn-vcons, inpaints frame-wise using Cxtattn, then reduces flickering with the blind video consistency method of Lai \etal~\cite{Lai2018}. The second phase of Cxtattn-vcons takes the original video and the frame-wise translated video as inputs, which in this case correspond to the masked video (with a placeholder value in the masked region) and the frame-wise inpainted video, respectively. The final method, VINet~\cite{Kim2019}, is a state-of-the-art DNN specifically designed and trained for video inpainting; we use their publicly-available inference code in our experiments. Among these methods, VINet is the only one tasked with inpainting videos at training time, and thus has an advantage over the other methods from having observed natural motion in masked videos.

\begin{figure}
  \centering
  \includegraphics[width=\linewidth]{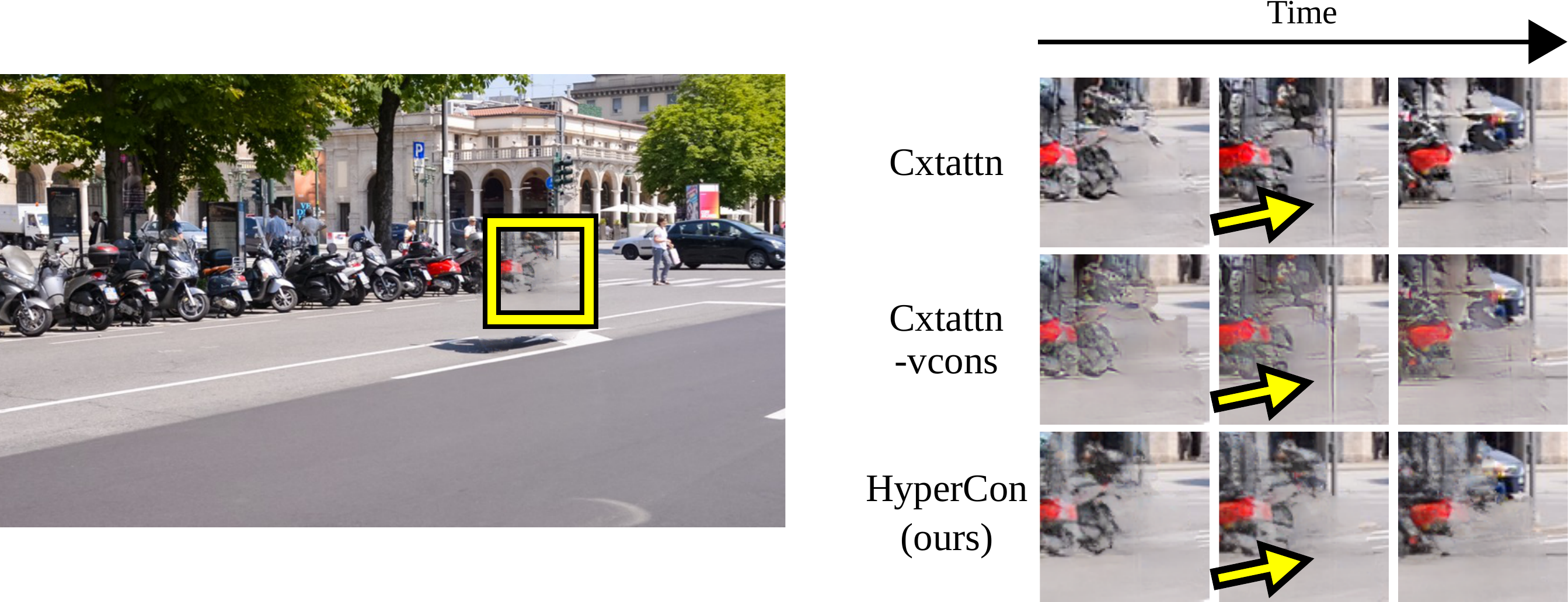}
  \caption{HyperCon substantially reduces flickering compared to the other image-to-video model transfer baselines Cxtattn and Cxtattn-vcons. Here, the pole appears for just one frame with the baselines but not with HyperCon.}
  \label{fig:flickering}
\end{figure}

\begin{figure}
  \centering
  \begin{subfigure}{\linewidth}
    \centering
    \begin{subfigure}{0.49\linewidth}
      \includegraphics[width=\linewidth]{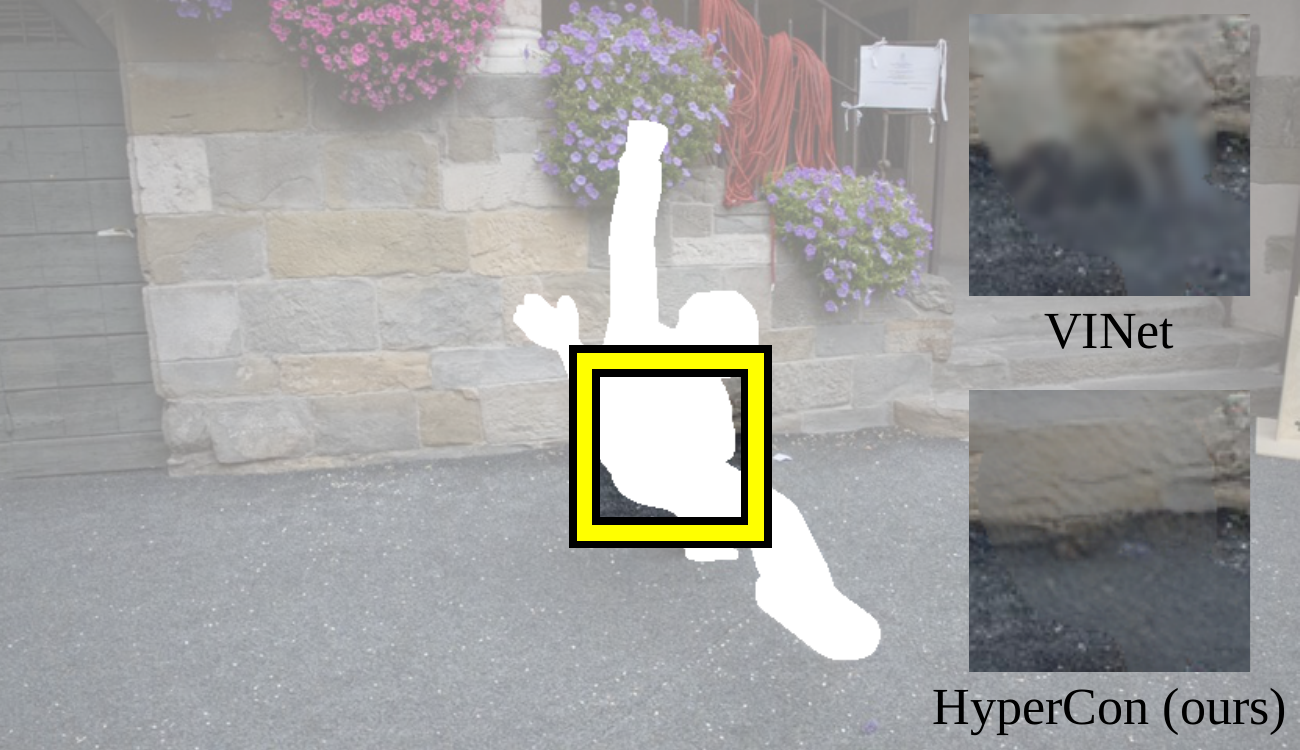}
      \caption{Boundary distortion}
      \label{fig:structure}
    \end{subfigure}
    \hfill
    \begin{subfigure}{0.49\linewidth}
      \includegraphics[width=\linewidth]{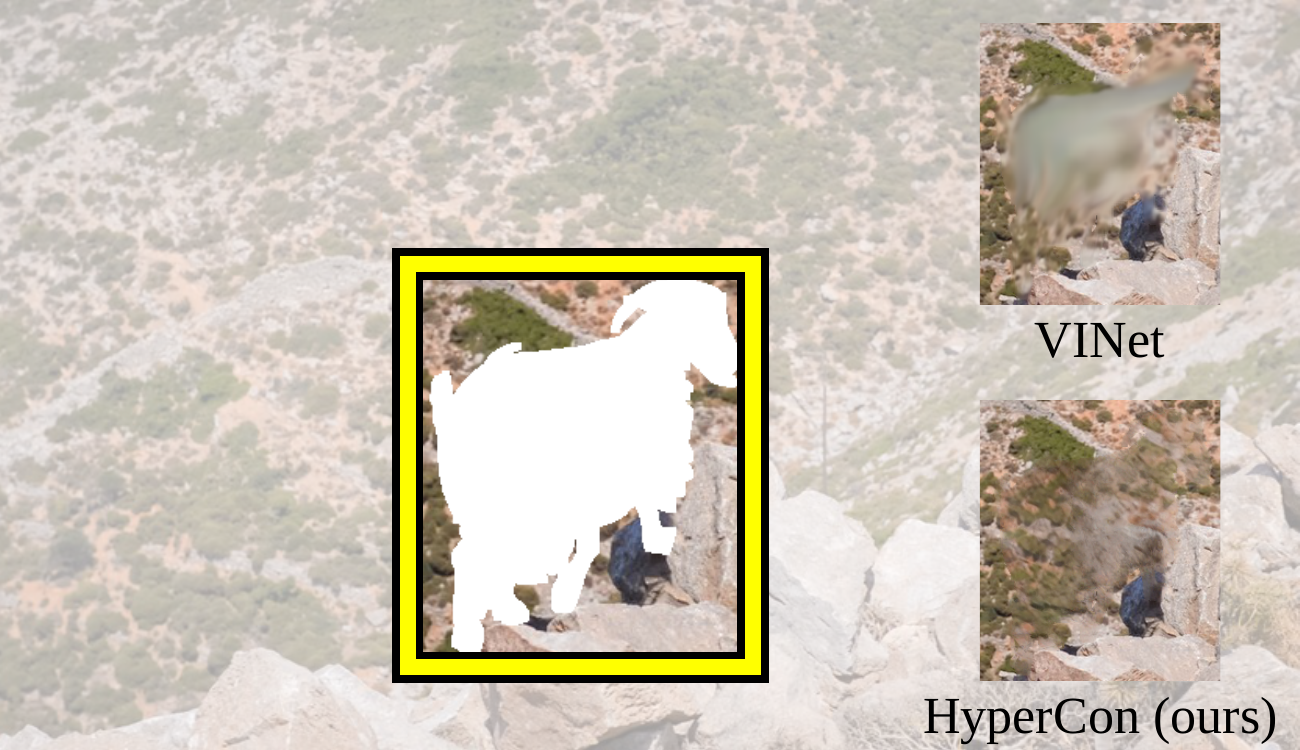}
      \caption{Texture comparison}
      \label{fig:vinet-textureless}
    \end{subfigure}
  \end{subfigure}
  \caption{Qualitative comparisons between VINet~\cite{Kim2019} and HyperCon (ours) for video inpainting. (a) VINet distorts the boundary of the masked wall, whereas HyperCon successfully recovers it. (b) VINet predicts textureless regions instead of realistic textures; in contrast, HyperCon produces sensible results due to its better generalization performance.}
\end{figure}

In Table~\ref{tab:swr-sor-quantitative}, we report quantitative results on our test videos for the inpainting task. Across the three test sets, HyperCon-mean obtains the lowest warping error, indicating that its predictions are most consistent with the motion of the ground-truth video (since warping error effectively checks against the estimated flow of the ground-truth video). Between HyperCon-mean and HyperCon-median, the latter generally scores slightly worse in $E_{\textrm{warp}}$ but better in the reconstruction and realism metrics; we suspect that the median filter produces sharper predictions that are more in line with real images, but are harder to match via optical flow (when computing $E_{\textrm{warp}}$) due to a wider variety of color intensities. Both HyperCon models generally outperform Cxtattn and Cxtattn-vcons across the evaluated datasets and metrics, suggesting that HyperCon is the most reliable method among image-to-video model transfer techniques. Impressively, HyperCon outperforms VINet by a substantial margin in all cases except on DAVIS 2019 under VFID \textit{despite not having seen a single masked video at training time}; this highlights the potential of transferring frame-wise models to video.

Next, we provide a qualitative analysis in the context of real-world object removal for DAVIS 2017 training/validation videos. Comparing our HyperCon method to Cxtattn and Cxtattn-vcons, we found several cases where the baseline methods produce flickering artifacts while ours does not (\eg, the pole in Figure~\ref{fig:flickering}). Additionally, due to the darkening behavior mentioned in Section~\ref{sec:comparison-to-prior-sota-style}, Cxtattn-vcons generates darkened predictions that do not blend in with the surrounding region as convincingly as those from HyperCon. Furthermore, HyperCon produces the fewest ``checkerboard artifacts''~\cite{odena2016} since they are temporally unstable and thus get filtered out by HyperCon during aggregation. We compare checkerboard artifacts and the darkening behavior in the supplementary materials.

We conclude our analysis by contrasting HyperCon with VINet~\cite{Kim2019}, the only evaluated method to have seen masked videos during training. We highlight two systematic failures of VINet that HyperCon overcomes: (i) boundary distortion and (ii) textureless prediction. Regarding the first failure, VINet often corrupts its inpainting result over time by incorrectly warping the inpainted structure. For example, in Figure~\ref{fig:structure}, VINet distorts the outline of the wall due to the continuous occlusion in that region; meanwhile, HyperCon successfully hallucinates the rigid wall boundary. As for the second issue, VINet sometimes initializes the inpainted region with a textureless prediction and fails to populate it with realistic texture throughout the video. In Figure~\ref{fig:vinet-textureless}, we compare this behavior with HyperCon, which generates sensible background textures. We posit that HyperCon is better able to hallucinate missing details because it has seen substantially more scenes than VINet during training (\ie, millions of images versus thousands of videos).

\section{Discussion and Conclusion}

In terms of weaknesses, HyperCon does not enforce long-range temporal dependencies beyond the aggregation dilation rate, so it cannot commit to stylization or inpainting choices for the entirety of extremely long videos. In addition, as with other image-to-video model transfer methods, HyperCon is subject to egregious errors of the frame-wise model, which can propagate downstream in certain cases. Integrating the three steps of HyperCon into one trainable, parameter-efficient model could help alleviate these issues by sharing more information between steps and enabling more input frames per unit of memory. Despite these pitfalls, HyperCon is still a promising image-to-video model transfer method for video-to-video translation tasks as seen by its strength in two widely different tasks. We hope that it helps pave the way for further progress in this direction.

\section*{Acknowledgements}

\noindent This work was completed in part while Ryan Szeto was an intern at Samsung Semiconductor, Inc. and was sponsored by DARPA and AFRL under agreement FA8750-17-2-0112, and the U.S. Department of Commerce, NIST under financial assistance award 60NANB17D191. The U.S. Government is authorized to reproduce and distribute reprints for Governmental purposes notwithstanding any copyright notation thereon. This work reflects the opinions and conclusions of its authors, but not necessarily its funding agents.


\newpage
\onecolumn

\vspace*{0.5cm}
\begin{center}
  \Large \textbf{HyperCon: Image-To-Video Model Transfer for\\Video-To-Video Translation Tasks -- Supplementary Materials}
\end{center}
\vspace*{1.5cm}

\appendix
\section{Additional Results for Style Transfer}

\begin{figure}[h]
  \centering
  \begin{subfigure}{0.32\linewidth}
    \includegraphics[width=\linewidth]{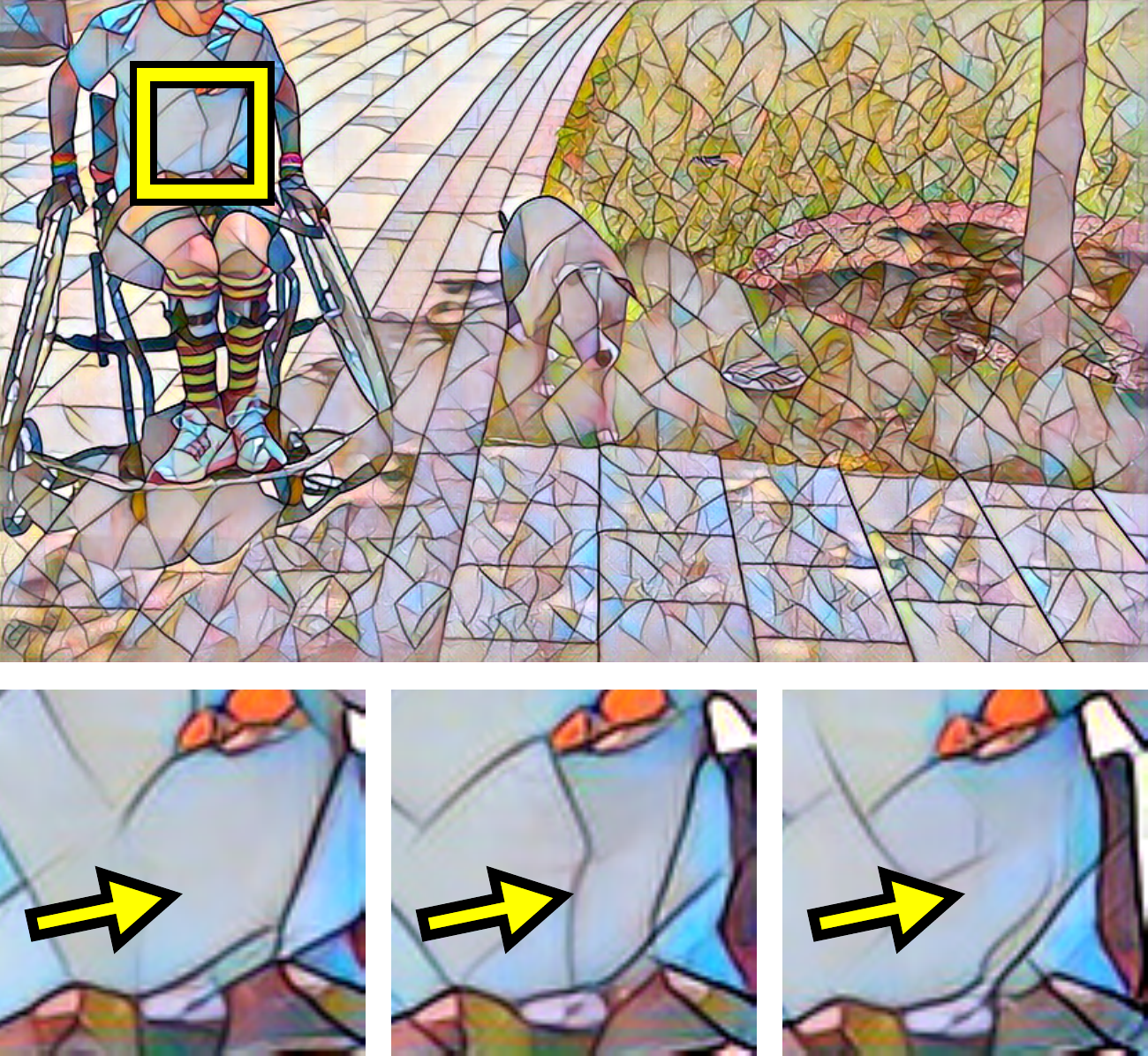}
    \caption{FST~\cite{Johnson2016}}
  \end{subfigure}
  \hfill
  \begin{subfigure}{0.32\linewidth}
    \includegraphics[width=\linewidth]{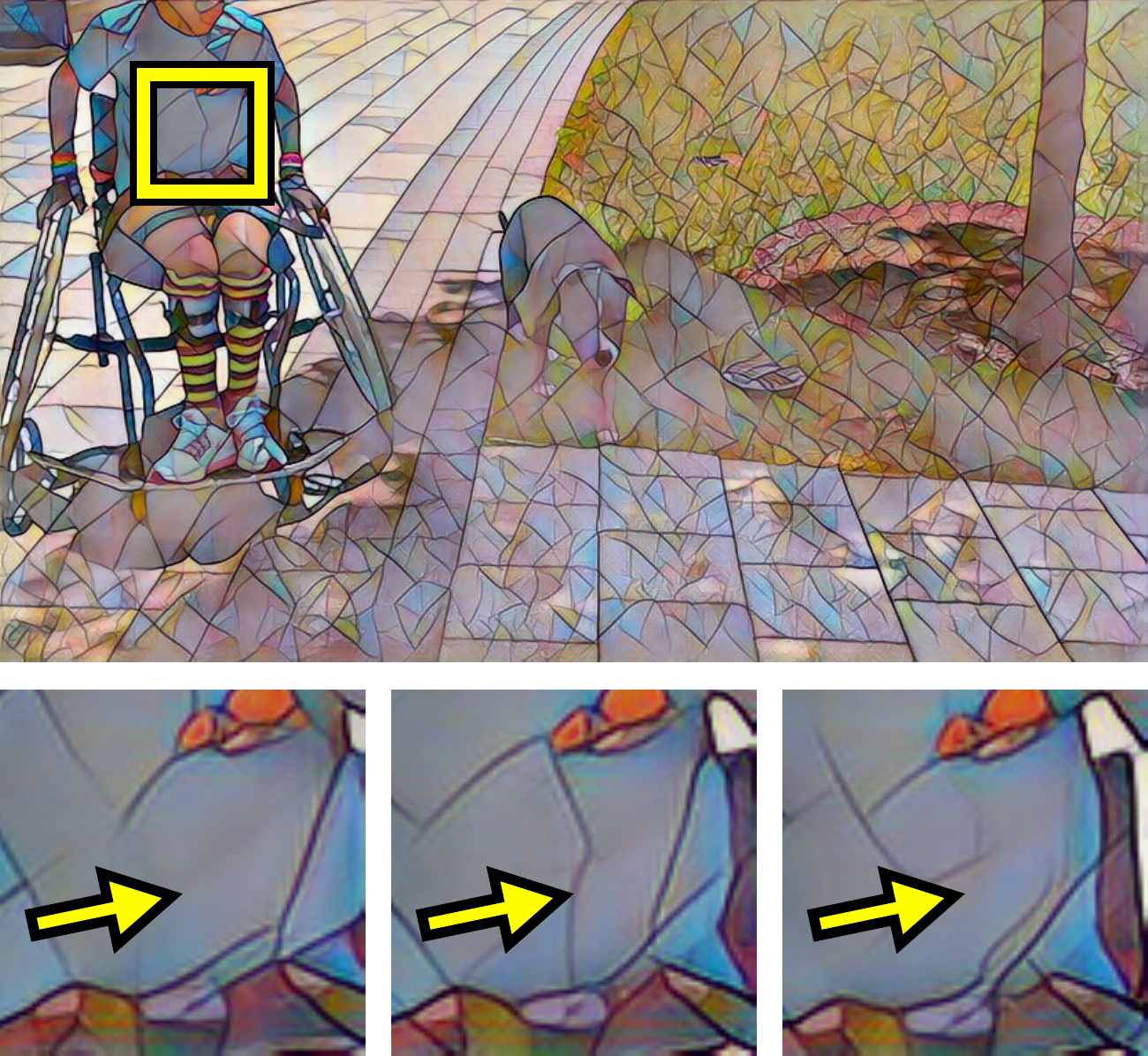}
    \caption{FST-vcons~\cite{Lai2018}}
  \end{subfigure}
  \hfill
  \begin{subfigure}{0.32\linewidth}
    \includegraphics[width=\linewidth]{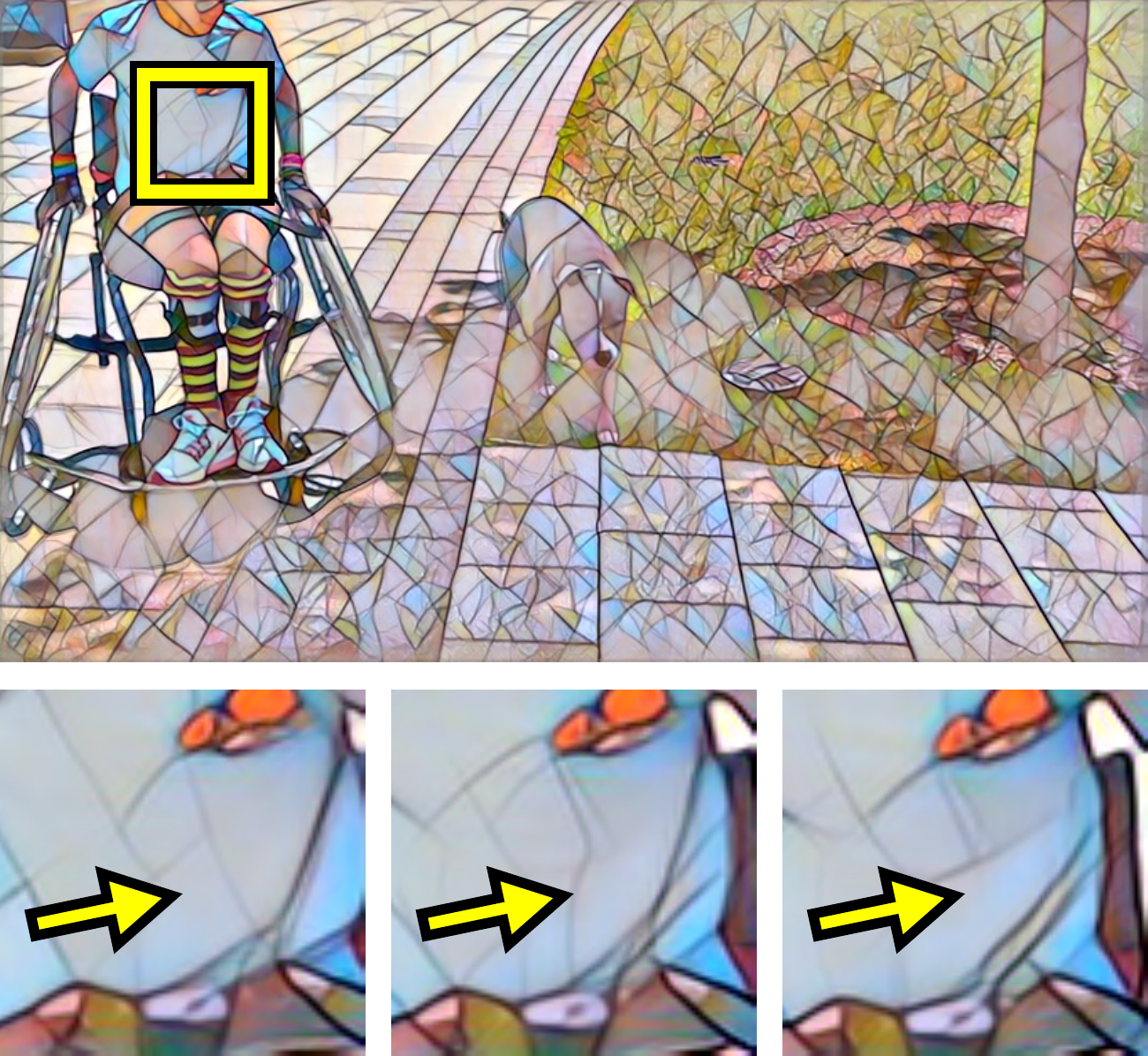}
    \caption{HyperCon (ours)}
  \end{subfigure}
  \caption{Comparison of flickering artifacts on the \textit{mosaic} style. The baselines (FST and FST-vcons) generate a vertical mark that appears for one frame, whereas HyperCon almost completely removes it.}
  \label{fig:baselines-style-supp-mosaic}
\end{figure}

\begin{figure}[h]
  \centering
  \begin{subfigure}{0.32\linewidth}
    \includegraphics[width=\linewidth]{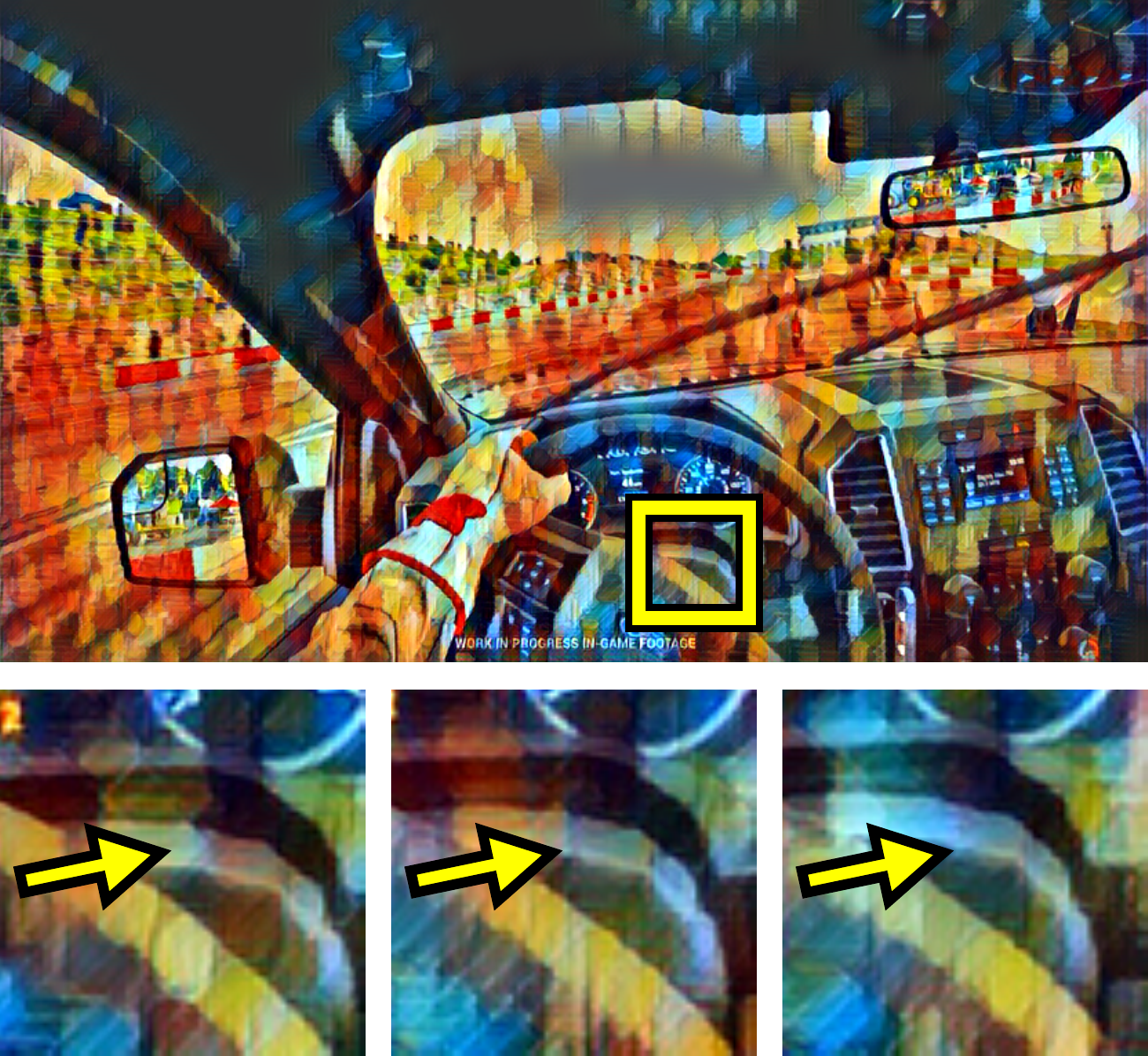}
    \caption{FST~\cite{Johnson2016}}
  \end{subfigure}
  \hfill
  \begin{subfigure}{0.32\linewidth}
    \includegraphics[width=\linewidth]{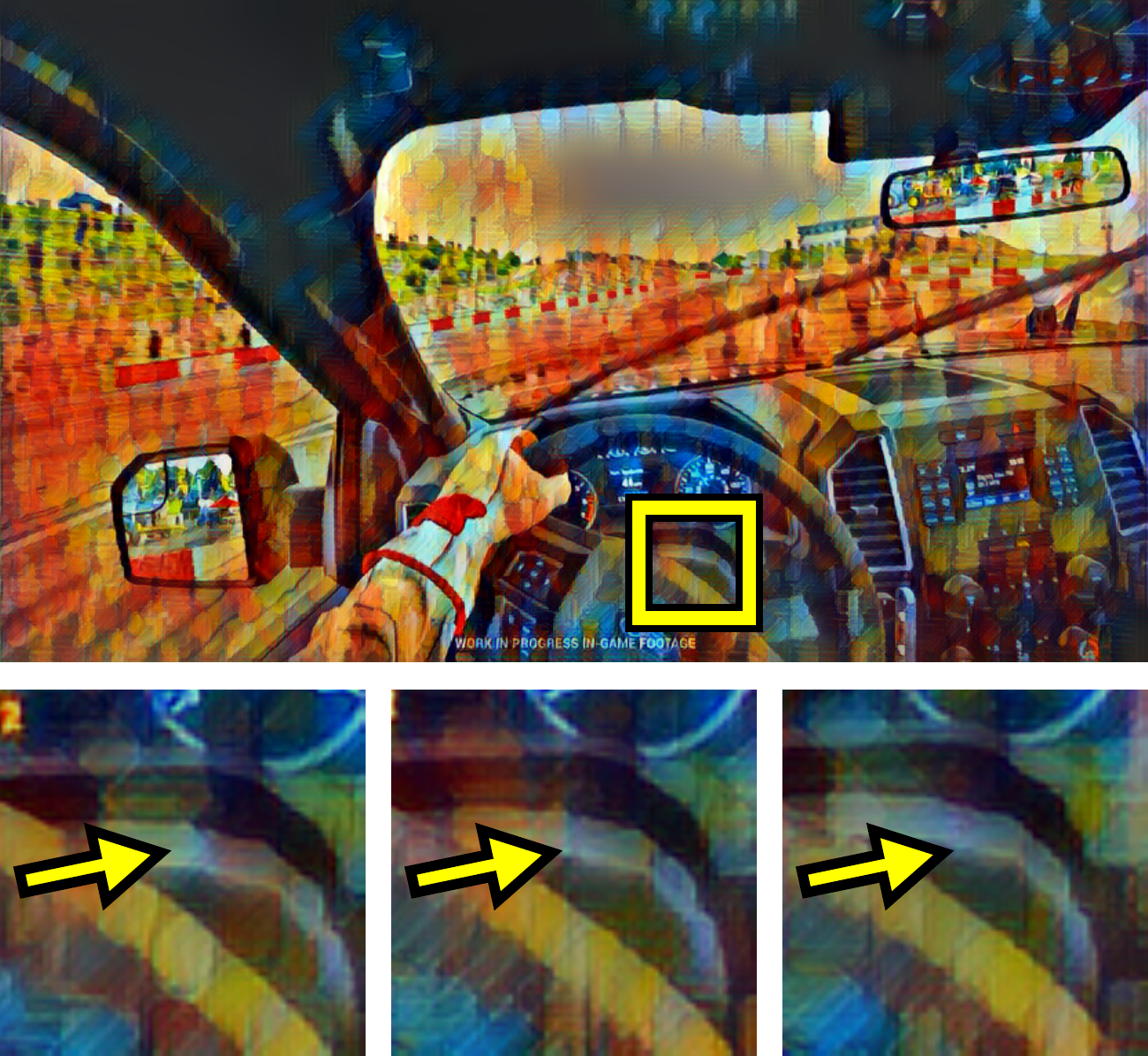}
    \caption{FST-vcons~\cite{Lai2018}}
  \end{subfigure}
  \hfill
  \begin{subfigure}{0.32\linewidth}
    \includegraphics[width=\linewidth]{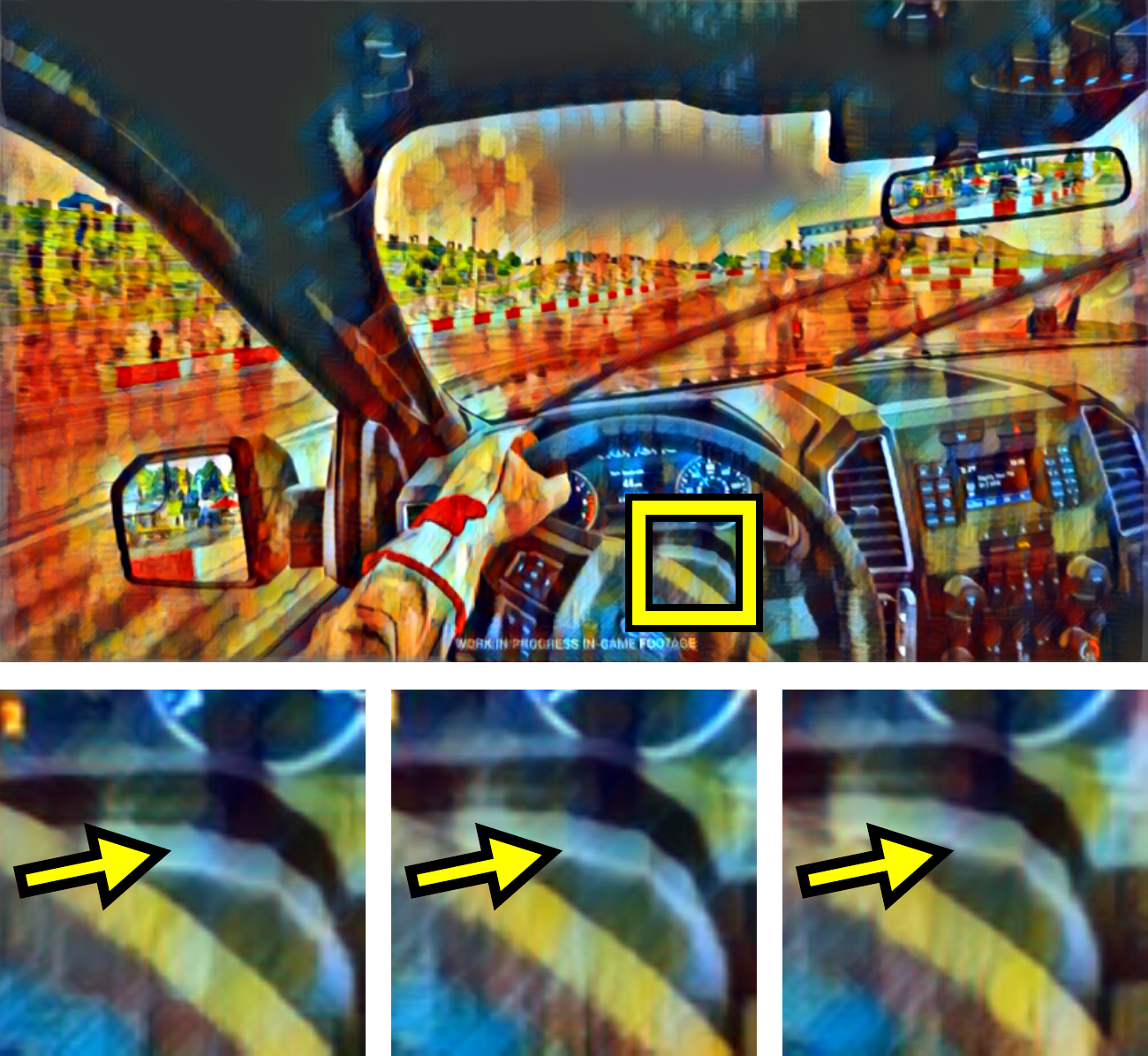}
    \caption{HyperCon (ours)}
  \end{subfigure}
  \caption{Comparison of flickering artifacts on the \textit{rain-princess} style. The baselines (FST and FST-vcons) frequently changes the colors within the indicated area, whereas HyperCon predicts a stable color across frames.}
  \label{fig:baselines-style-supp-rain-princess}
\end{figure}

In Figures~\ref{fig:baselines-style-supp-mosaic} and \ref{fig:baselines-style-supp-rain-princess}, we compare HyperCon's predictions to the baselines for additional style transfer examples. We see that HyperCon produces more temporally consistent predictions than the baselines.

\subsection{Ablative Analysis}

\begin{figure}
  \centering
  \begin{subfigure}{\linewidth}
    \centering
    \begin{subfigure}{0.49\linewidth}
      \includegraphics[width=\linewidth]{figs/ablation-style/fid-davis-rain-princess.pdf}
    \end{subfigure}
    \begin{subfigure}{0.49\linewidth}
      \includegraphics[width=\linewidth]{figs/ablation-style/err_warp-davis-rain-princess.pdf}
    \end{subfigure}
    \caption{\textit{rain-princess}, DAVIS train/val}
  \end{subfigure}
  \begin{subfigure}{\linewidth}
    \centering
    \begin{subfigure}{0.49\linewidth}
      \includegraphics[width=\linewidth]{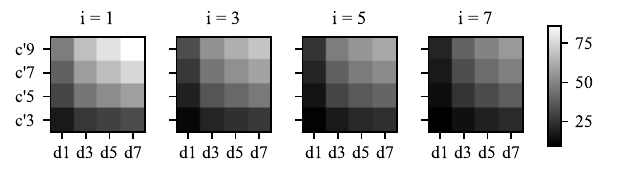}
    \end{subfigure}
    \begin{subfigure}{0.49\linewidth}
      \includegraphics[width=\linewidth]{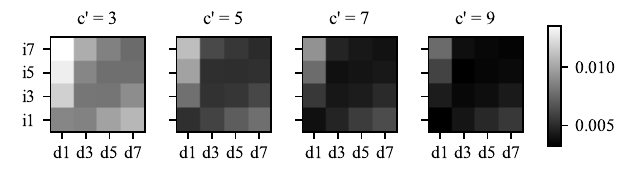}
    \end{subfigure}
    \caption{\textit{mosaic}, DAVIS train/val}
  \end{subfigure}
  \begin{subfigure}{\linewidth}
    \centering
    \begin{subfigure}{0.49\linewidth}
      \includegraphics[width=\linewidth]{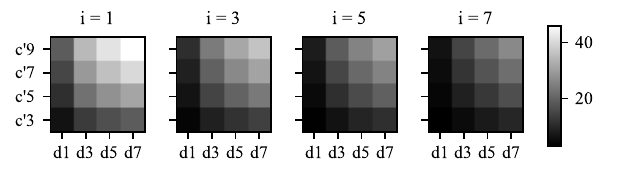}
    \end{subfigure}
    \begin{subfigure}{0.49\linewidth}
      \includegraphics[width=\linewidth]{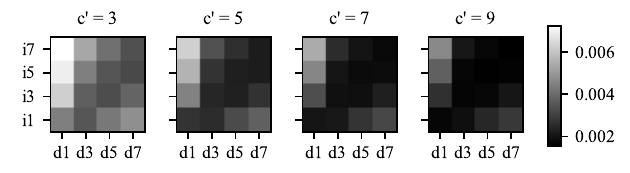}
    \end{subfigure}
    \caption{\textit{rain-princess}, ActivityNet val}
  \end{subfigure}
  \begin{subfigure}{\linewidth}
    \centering
    \begin{subfigure}{0.49\linewidth}
      \includegraphics[width=\linewidth]{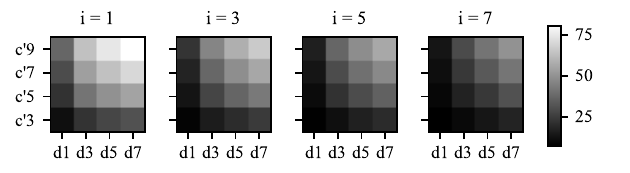}
    \end{subfigure}
    \begin{subfigure}{0.49\linewidth}
      \includegraphics[width=\linewidth]{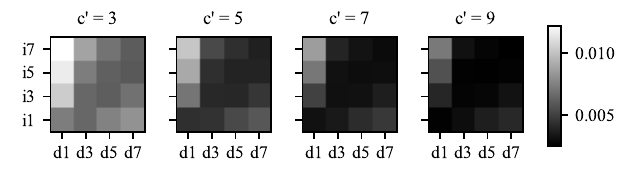}
    \end{subfigure}
    \caption{\textit{mosaic}, ActivityNet val}
  \end{subfigure}
  \caption{Ablative analysis plots. Left plots show FID (style adherence) and right plots show $E_{\textrm{warp}}$ (temporal consistency). Lower is better.}
  \label{fig:ablative-quantitative-supp}
\end{figure}

\begin{figure}
  \centering
  \includegraphics[width=0.7\linewidth]{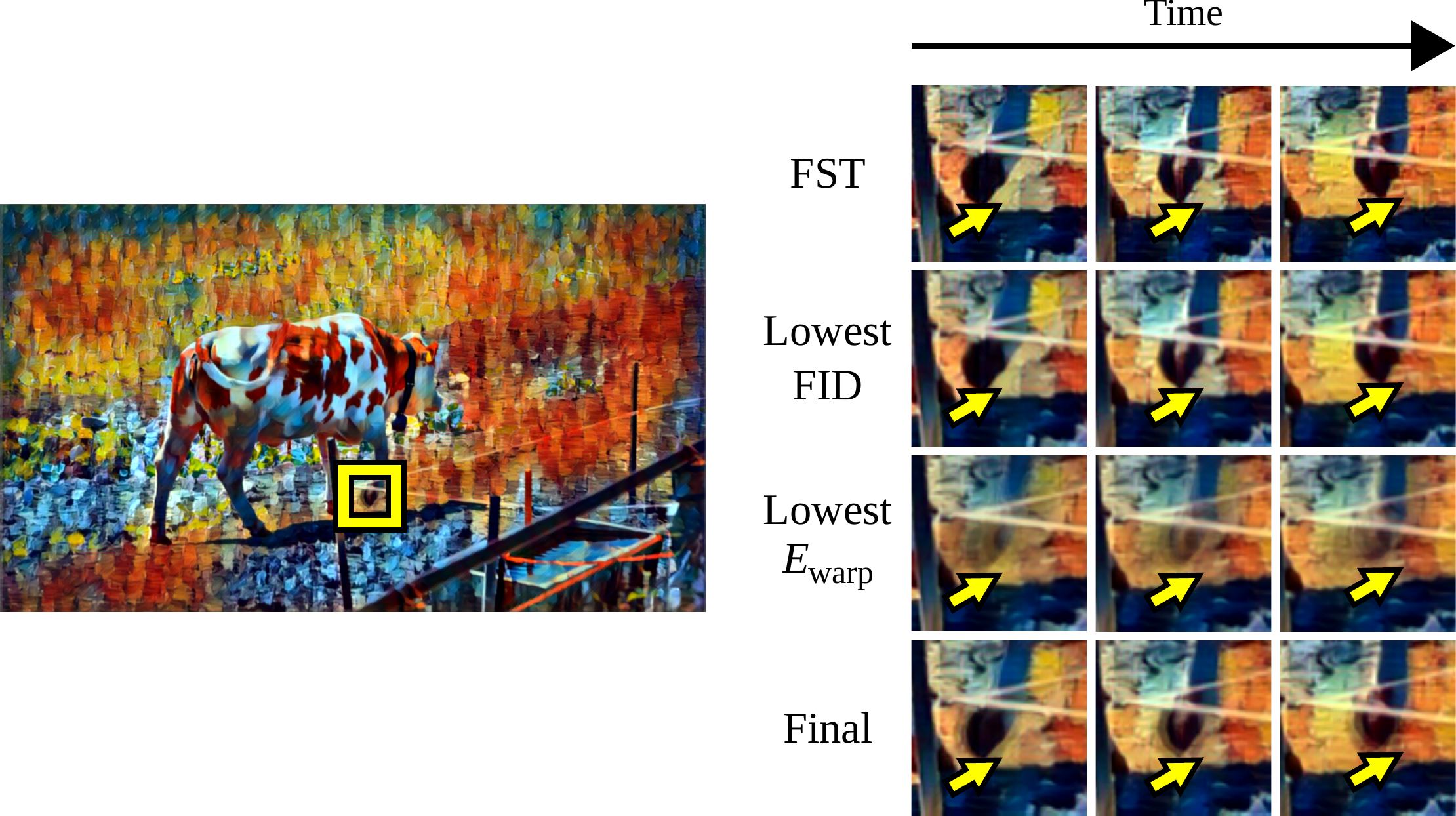}
  \caption{Qualitative comparison between frame-wise style transfer (FST) and ablative variants of HyperCon. The ``Lowest FID'' model reproduces flickering artifacts (\eg, the changing tone near the arrow), while the ``Lowest $E_{\textrm{warp}}$'' model overly blurs predictions. Our final model adheres to the intended style without overly blurring predictions.}
  \label{fig:ablative-qualitative}
\end{figure}

Figure~\ref{fig:ablative-quantitative-supp} includes quantitative plots for our ablative analysis for all styles and validation sets; we observe similar trends across all styles and validation sets. Turning to qualitative results in Figure~\ref{fig:ablative-qualitative}, we confirm that a low FID indicates strong style adherence and that a low $E_{\textrm{warp}}$ indicates strong temporal consistency. However, strictly minimizing one or the other does not yield the most visually satisfying results. For instance, the hyperparameters that minimize FID exhibit the same flickering artifacts as frame-wise style transfer (FST)---observe that the region next to the cow's foot suddenly changes from blue to orange in the final frame for both FST and the lowest FID model. Meanwhile, the hyperparameters that minimize $E_{\textrm{warp}}$ yield blurry predictions, which is the result of overly smearing several intermediate predictions across frames. Our final model produces consistent tones without overblurring, indicating that our selection strategy sensibly compromises between the two objectives.

\subsection{Style Transfer Survey Design}

To gauge the quality of our style transfer results, we have conducted a human evaluation on Amazon Mechanical Turk\footnote{\url{https://www.mturk.com/}} (AMT). The motivation behind this study is primarily to ensure that the quantitative metrics that we use in the main paper correlate with human judgement at scale.

\begin{figure}
  \centering
  \begin{subfigure}{\linewidth}
    \centering
    \includegraphics[width=\linewidth]{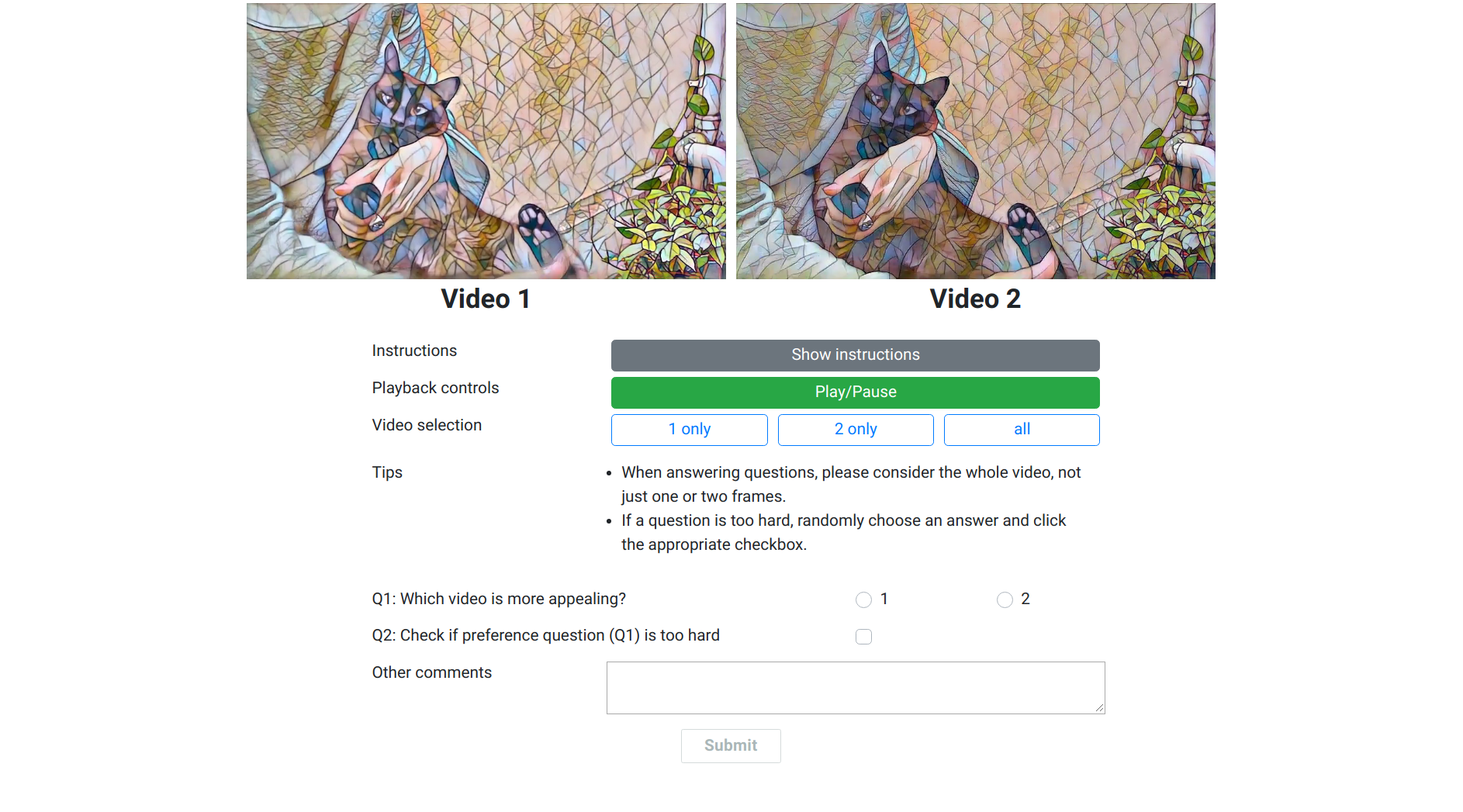}
    \caption{Preferred}
    \label{fig:survey-preferred}
  \end{subfigure}
  \begin{subfigure}{\linewidth}
    \centering
    \includegraphics[width=\linewidth]{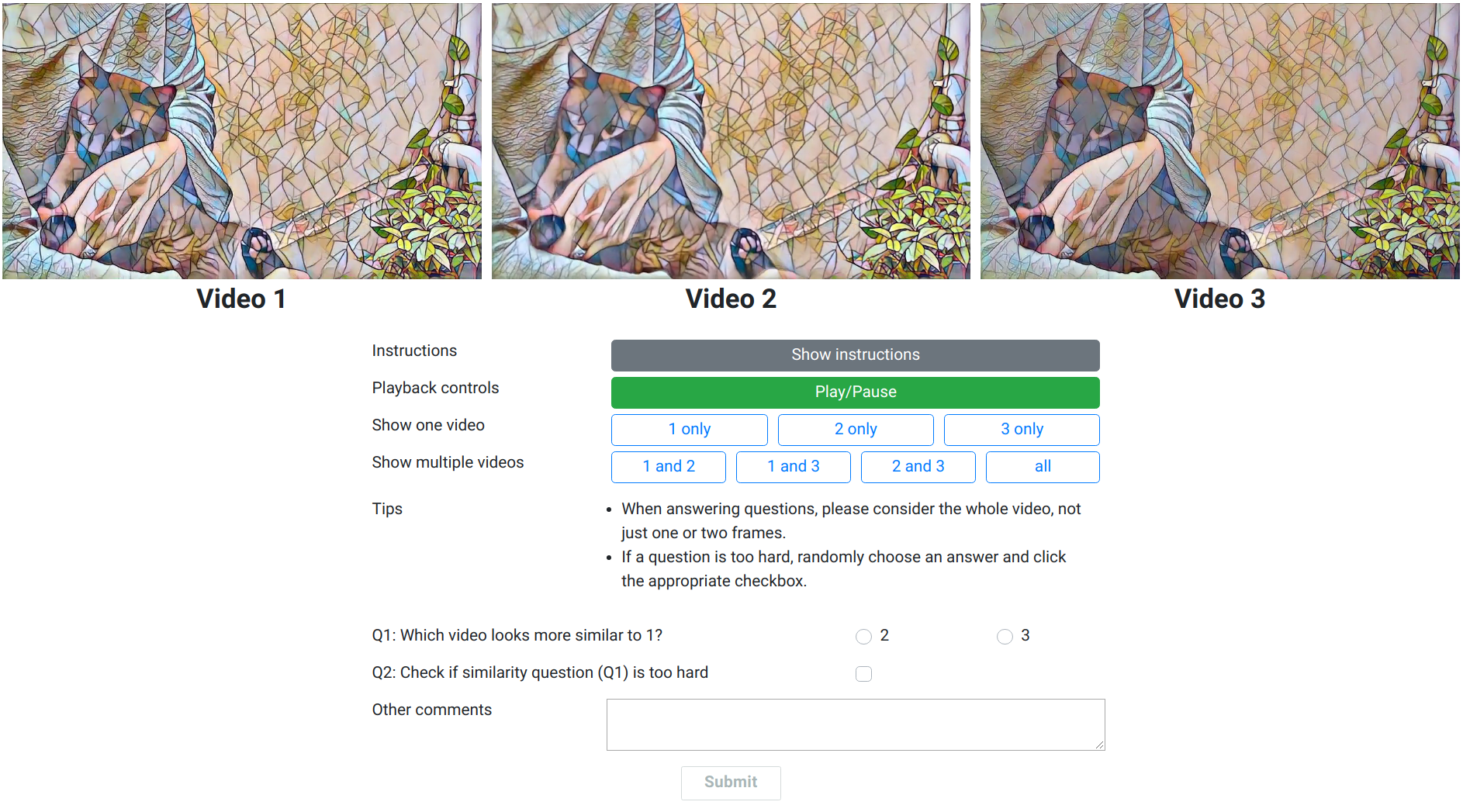}
    \caption{Style adherence}
    \label{fig:survey-style-adherence}
  \end{subfigure}
  \caption{Examples of the interfaces used for the ``preferred'' and ``style adherence'' surveys.}
  \label{fig:survey}
\end{figure}

For our study, we developed custom web interfaces shown in Figure~\ref{fig:survey}. The displayed videos are synchronized at all times, and playback can be toggled by pressing the ``Play/Pause'' button. Users can choose to display one or multiple videos at a time; if one video is selected, it is centered at the top of the page, and if multiple videos are selected, any non-selected video is replaced with a black frame. Playback does not reset when different videos are selected, so users can effectively perform A/B-style viewing without losing their place in the video.

Using these interfaces, we asked subjects two questions:
\begin{itemize}
  \item ``Which video is more appealing?'' (preferred)
  \item ``Which video looks more to 1?'' (style adherence)
\end{itemize}
For the ``preferred'' question, we intentionally omitted instructions on what exactly makes a video more appealing since such qualities are inherently ambiguous. As for the ``style adherence'' question, we framed it in terms of comparing the pairs of videos under evaluation to the frame-wise stylized video as reference; we determined this to be the most concrete, unambiguous way to define the intended video stylization for subjects.

For each combination of style and source video, we randomly selected which method (between HyperCon and the FST-vcons baseline) would appear as the first and second video under evaluation (\ie, videos 1 and 2 in Figure~\ref{fig:survey-preferred} and videos 2 and 3 in Figure~\ref{fig:survey-style-adherence}). This forces subjects to watch all videos carefully when answering each question. For the ``style adherence'' question, the frame-wise stylized video (FST) always appeared as video 1.

We also provided a checkbox to allow the subject to indicate that a question was difficult. We had considered an alternative survey formulation that only asked one question, but offered three responses (\eg, ``1'', ``2'', or ``neither'' for the ``preferred'' survey), but opted for the two-question approach to ensure that rejecting the null hypothesis would lead to an interpretable result. We observed that the checkbox was selected for 5.08\% of all responses to the ``preferred'' question and 6.61\% of all responses to the ``style adherence'' question, indicating that the two methods were clearly distinctive in most cases.

We paid \$0.10 USD per task, which equates to \$12 USD/hour if each survey takes 30 seconds to answer. A total of about 150 subjects participated in our experiments. Surveys were generated as separate Human Intelligence Tasks (HITs), so it is not necessarily the case that any given subject saw both questions for any or all videos/styles.

In terms of filtering subjects, we did not utilize any demographic-based filters; instead, we created a qualification test utilizing an interface equivalent to Figure~\ref{fig:survey-style-adherence}. The videos used in this test were derived from a source video not included in our DAVIS and ActivityNet test videos, and had a simple variable darkening filter applied to each frame (\ie, multiplying all RGB values by some constant). Examples of the effects we applied include darkening all frames by the same amount, darkening each frame by a random amount, and not darkening any frame. For each video triplet, we always included one of the possible answers as the reference; this effectively forced subjects to match the reference video with one of the other two. We also ensured that the odd video out was blatantly different to make the qualification test straightforward. We generated 10 triplets of qualification test videos, and required subjects to answer correctly for at least 8 of them before working on our tasks. For our final survey, we solicited 5 responses for each combination of source video, style, and question.

\subsection{Detailed Style Transfer Survey Results}

\begin{table}
  \begin{subfigure}{\linewidth}
    \centering
    \scalebox{0.7}{
      \begin{tabular}{c|cc|cc}
        \hline
         & \multicolumn{2}{c|}{mosaic} & \multicolumn{2}{c}{rain-princess} \\
        \cline{2-5}
         & Preferred & Style adherence & Preferred & Style adherence \\
        \hline
        FST-vcons~\cite{Lai2018} & 42.7\% & 20.9\% & 44.5\% & 23.2\% \\
        HyperCon (ours) & \textbf{57.3\%} & \textbf{79.1\%} & \textbf{55.5\%} & \textbf{76.8\%} \\
        \hline
        p-value & $5.53 \times 10^{-7}$ & 0* & $1.54 \times 10^{-4}$ & 0* \\
        \hline
      \end{tabular}
    }
    \caption{}
    \label{tab:amt-style}
  \end{subfigure}

  \begin{subfigure}{\linewidth}
    \vspace{6pt}
    \centering
    \scalebox{0.7}{
      \begin{tabular}{c|cc|cc|cc|cc|cc|cc}
        \hline
         & \multicolumn{4}{c|}{DAVIS 2017 (60 source videos)} & \multicolumn{4}{c|}{DAVIS 2019 (60 source videos)} & \multicolumn{4}{c}{ActivityNet (116 source videos)} \\
        \cline{2-13}
         & \multicolumn{2}{c|}{mosaic} & \multicolumn{2}{c|}{rain-princess} & \multicolumn{2}{c|}{mosaic} & \multicolumn{2}{c|}{rain-princess} & \multicolumn{2}{c|}{mosaic} & \multicolumn{2}{c}{rain-princess} \\
        \hline
        Method & Preferred & \begin{tabular}[c]{@{}c@{}}Style\\adherence\end{tabular} & Preferred & \begin{tabular}[c]{@{}c@{}}Style\\adherence\end{tabular} & Preferred & \begin{tabular}[c]{@{}c@{}}Style\\adherence\end{tabular} & Preferred & \begin{tabular}[c]{@{}c@{}}Style\\adherence\end{tabular} & Preferred & \begin{tabular}[c]{@{}c@{}}Style\\adherence\end{tabular} & Preferred & \begin{tabular}[c]{@{}c@{}}Style\\adherence\end{tabular} \\
        \hline
        FST-vcons~\cite{Lai2018} & 137 & 45 & 128 & 68 & 133 & 43 & 147 & 63 & 234 & 159 & 250 & 143 \\
        HyperCon (ours) & \textbf{163} & \textbf{255} & \textbf{172} & \textbf{232} & \textbf{167} & \textbf{257} & \textbf{153} & \textbf{237} & \textbf{346} & \textbf{421} & \textbf{330} & \textbf{437} \\
        \hline
        Total & 300 & 300 & 300 & 300 & 300 & 300 & 300 & 300 & 580 & 580 & 580 & 580 \\
        \hline
      \end{tabular}
    }
    \caption{}
    \label{tab:amt-style-detailed}
  \end{subfigure}
  \caption{Human evaluation of style transfer quality. (a) For each style, we list how often subjects favorably select each method across all videos of that style, as well as the p-value of the corresponding $\chi^2$ test. 0* indicates a p-value less than $1\times10^{-10}$. In all cases, subjects select HyperCon (ours) significantly more often than FST-vcons~\cite{Lai2018}. (b) A detailed breakdown of responses for each style and dataset.}
\end{table}

We provide an aggregate view of our human evaluation results in Table~\ref{tab:amt-style}. For the ``preferred'' question, subjects select HyperCon more often than FST-vcons, enough to reject the null hypothesis that subjects select each method with equal probability---in short, they prefer our predictions over those of the baseline. Furthermore, for the ``style adherence'' question, which asks for the video that is more similar to the reference style video, subjects also select our HyperCon method significantly more often; this means that our method is better at preserving the intended style than FST-vcons. This matches the conclusion made from our method's lower FID scores, and indicates that FID correlates with video quality and style adherence. In Table~\ref{tab:amt-style-detailed}, we provide a detailed breakdown of responses for each question and dataset.

\section{Additional Qualitative Results for Inpainting}

\begin{figure}[H]
  \centering
  \begin{subfigure}[b]{0.32\linewidth}
    \includegraphics[width=\linewidth]{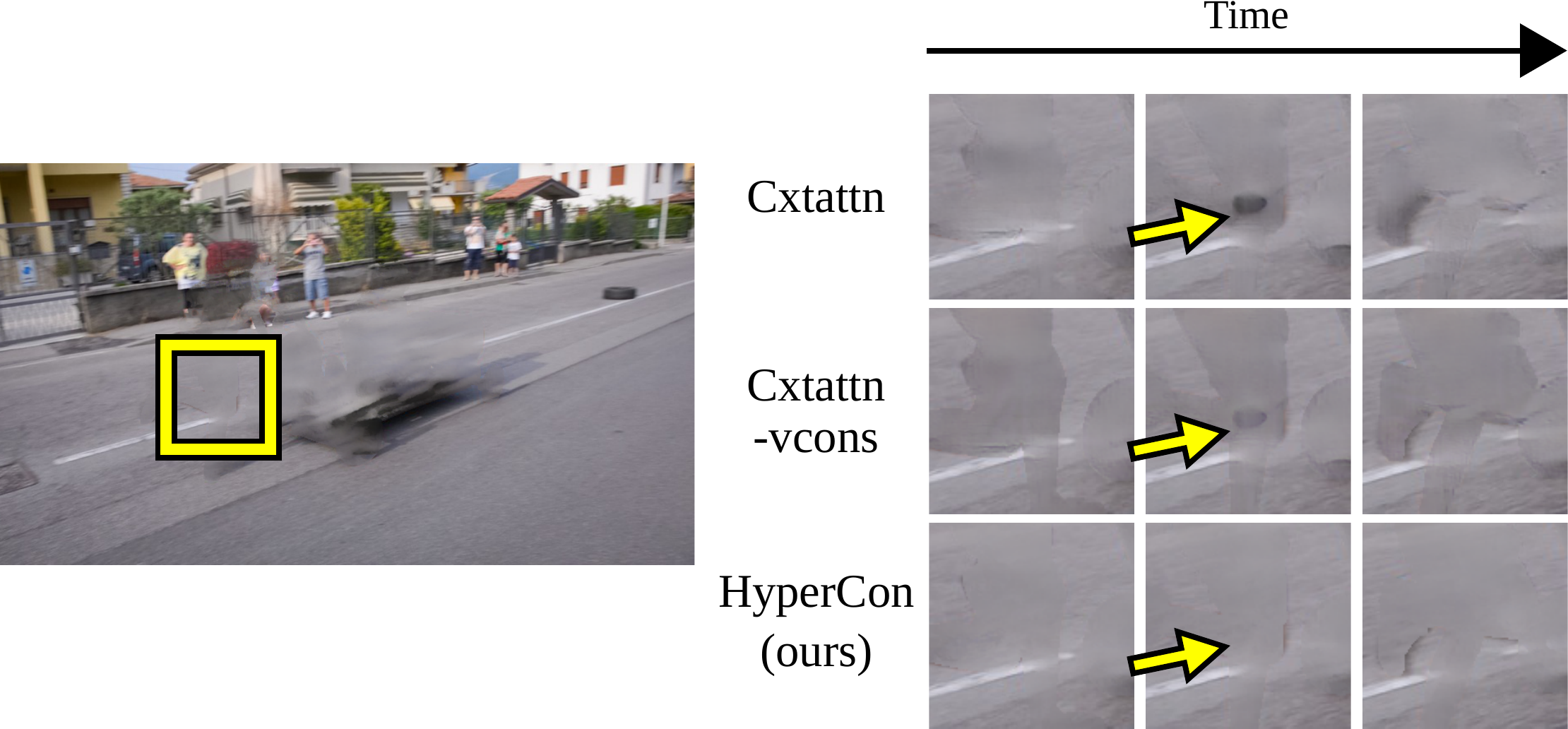}
    \caption{Flickering}
  \end{subfigure}
  \hfill
  \begin{subfigure}[b]{0.32\linewidth}
    \includegraphics[width=\linewidth]{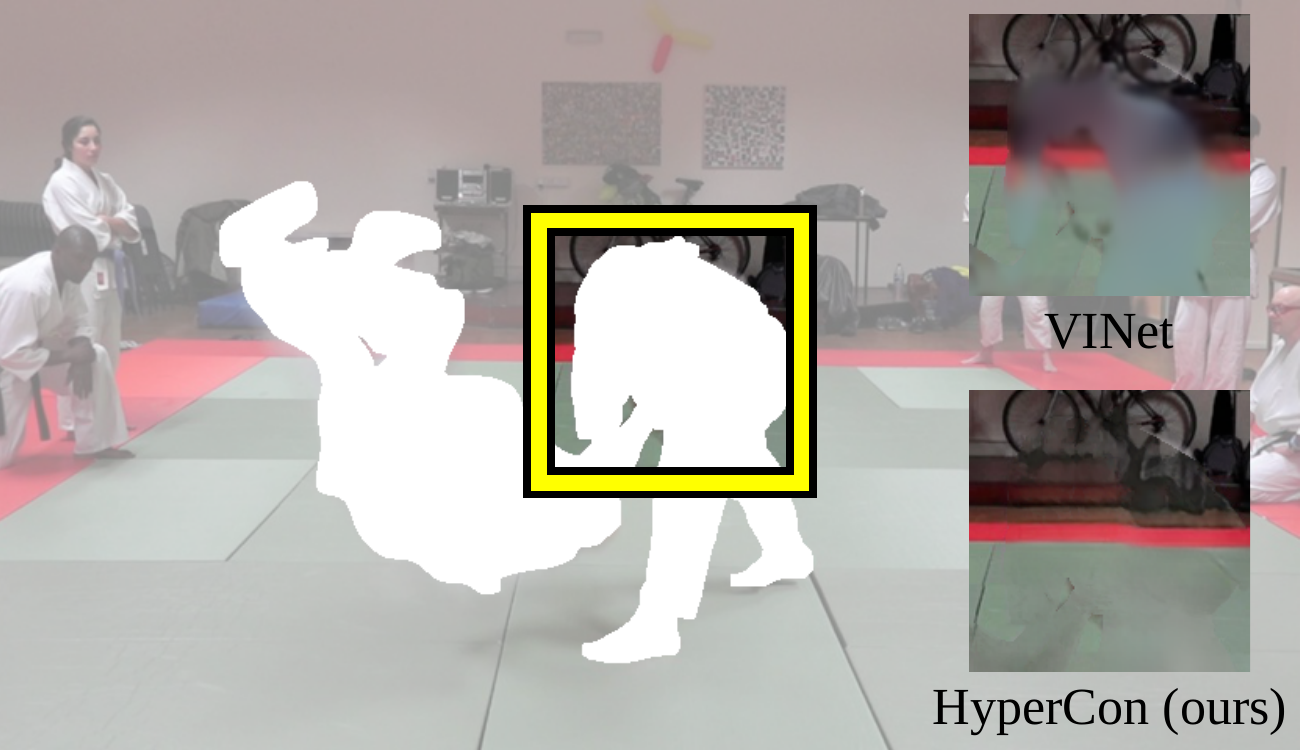}
    \caption{Boundary distortion}
  \end{subfigure}
  \hfill
  \begin{subfigure}[b]{0.32\linewidth}
    \includegraphics[width=\linewidth]{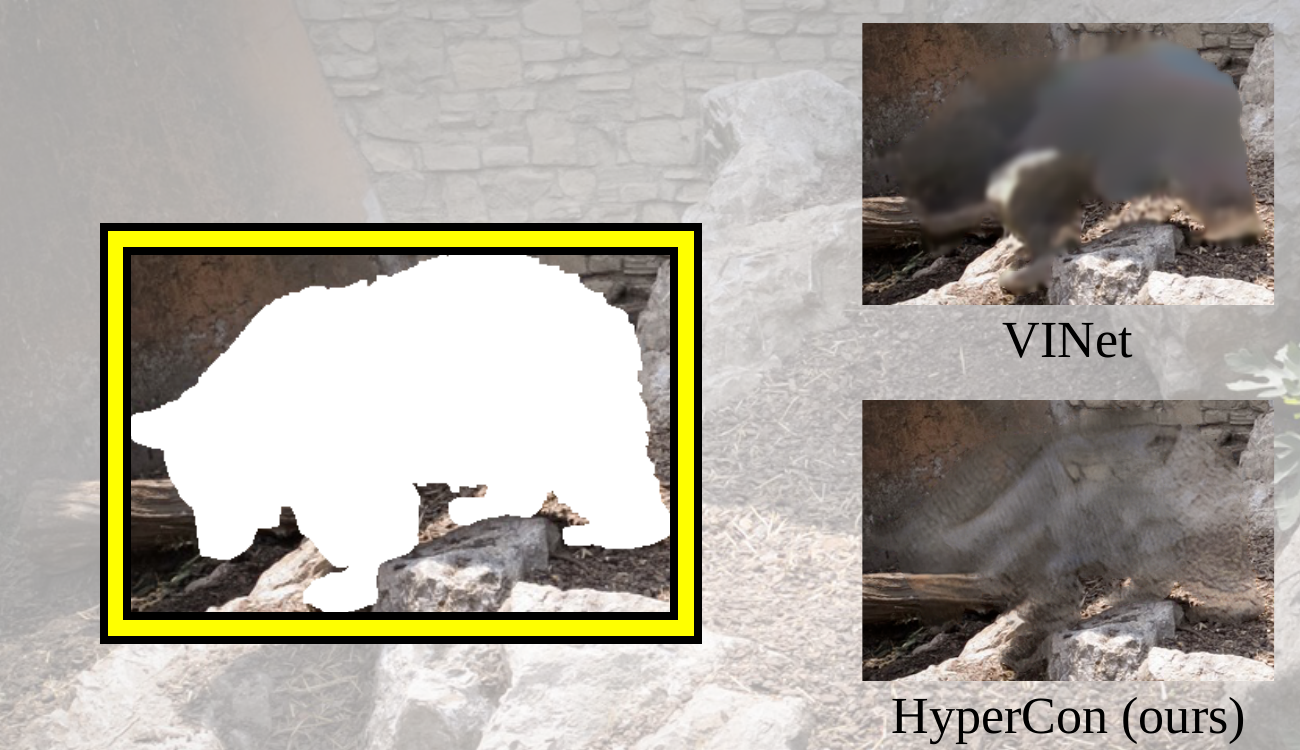}
    \caption{Texture comparison}
  \end{subfigure}
  \caption{Additional examples of flickering and boundary distortion effects from the baselines versus HyperCon for video inpainting. (a) HyperCon reduces artifacts better than the image-to-video model transfer baselines. The gray circle is apparent in the Cxtattn and Cxtattn-vcons prediction, but not in the HyperCon one. (b) VINet fails to connect the boundary of the mat in the background, whereas HyperCon successfully does connect the boundary. (c) VINet produces overly smooth textures that do not blend in well with the surrounding region, whereas HyperCon produces more realistic textures.}
  \label{fig:flickering-structure-supp}
\end{figure}

\begin{figure}[h]
  \centering
  \begin{subfigure}{0.6\linewidth}
    \centering
    \includegraphics[width=\linewidth]{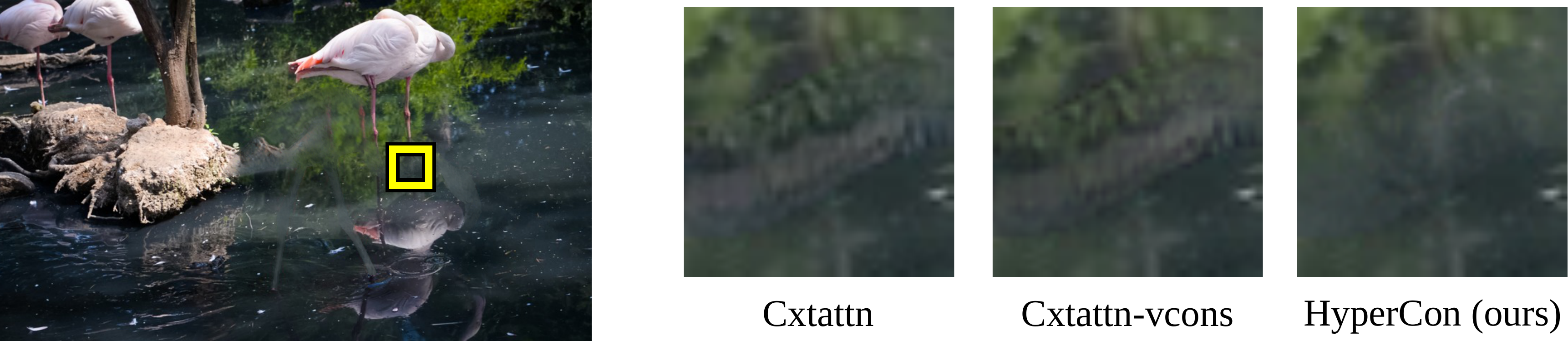}
    \caption{Flamingo}
  \end{subfigure}
  \begin{subfigure}{0.6\linewidth}
    \centering
    \includegraphics[width=\linewidth]{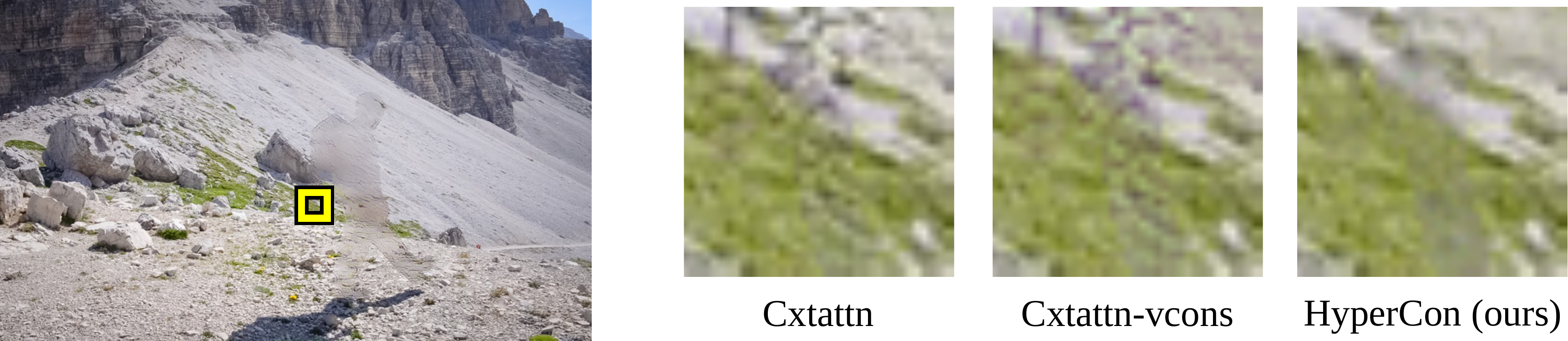}
    \caption{Hike}
  \end{subfigure}
  \caption{HyperCon generates fewer checkerboard artifacts than the baselines due to their instability across frames.}
  \label{fig:checkerboard}
\end{figure}

\begin{figure}[H]
  \centering
  \begin{subfigure}{0.24\linewidth}
    \centering
    \includegraphics[width=\linewidth]{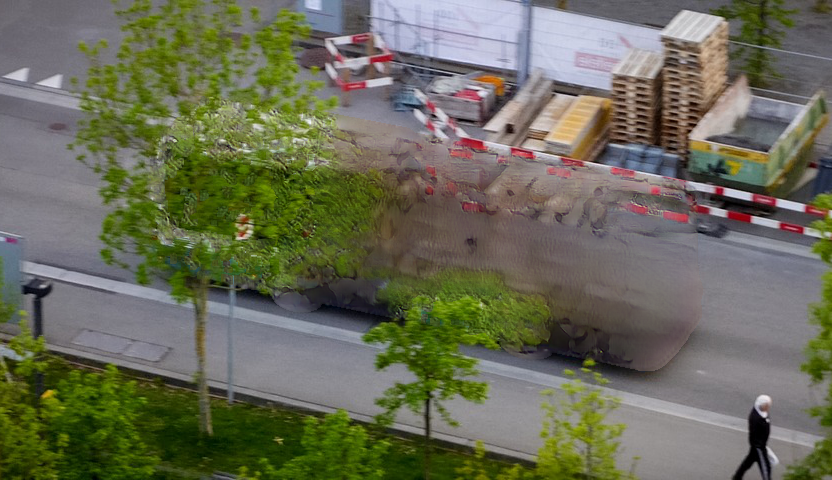}
    \caption{Cxtattn-vcons}
  \end{subfigure}
  \hfill
  \begin{subfigure}{0.24\linewidth}
    \centering
    \includegraphics[width=\linewidth]{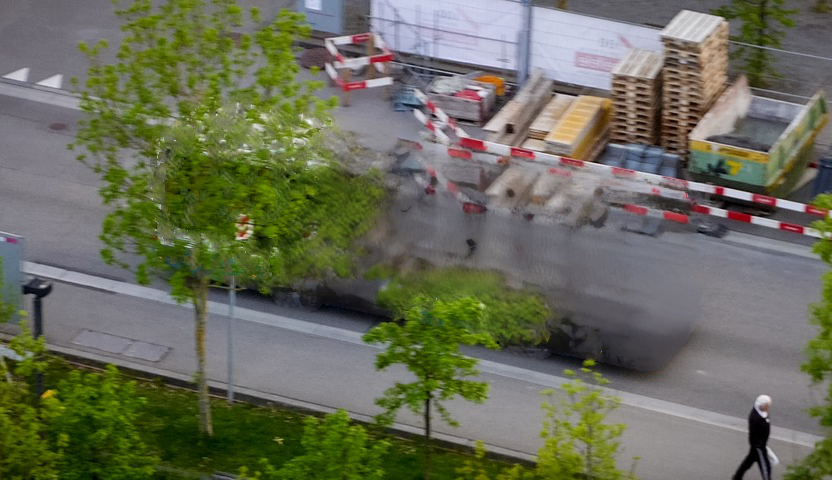}
    \caption{HyperCon (ours)}
  \end{subfigure}
  \hfill
  \begin{subfigure}{0.24\linewidth}
    \centering
    \includegraphics[width=\linewidth]{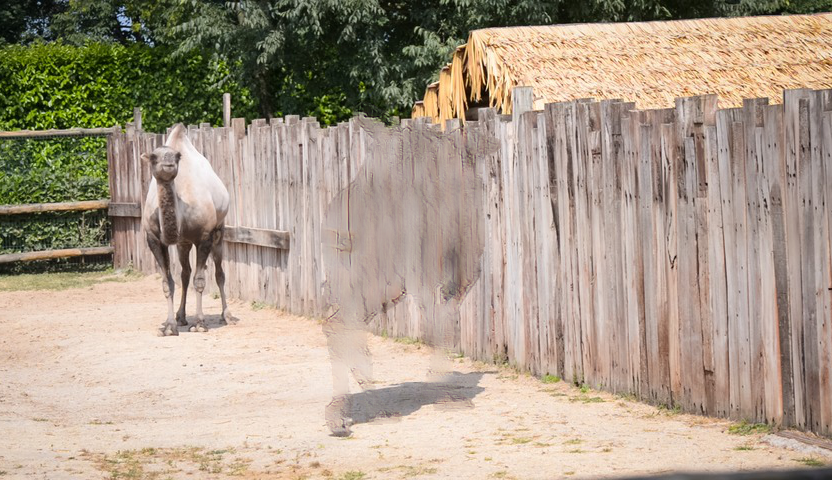}
    \caption{Cxtattn-vcons}
  \end{subfigure}
  \hfill
  \begin{subfigure}{0.24\linewidth}
    \centering
    \includegraphics[width=\linewidth]{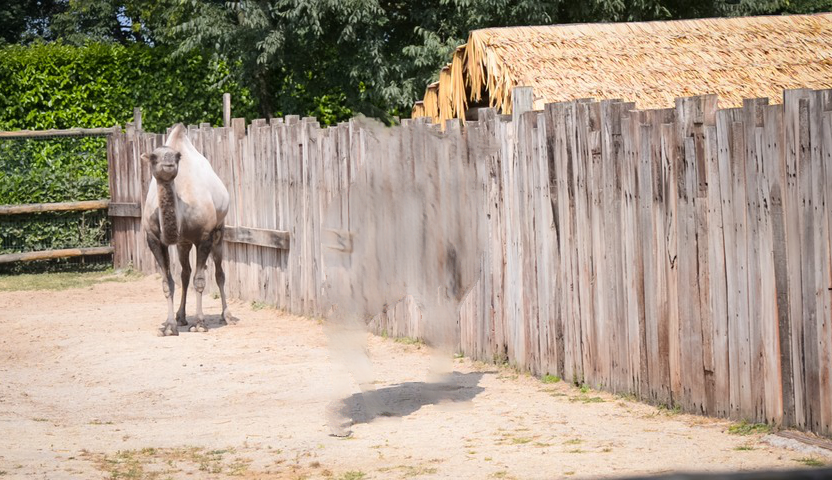}
    \caption{HyperCon (ours)}
  \end{subfigure}
  \caption{Cxtattn-vcons distorts the hue of the inpainted region; HyperCon does not. As a result, our HyperCon predictions blend in more convincingly with the surrounding area.}
  \label{fig:vcons-sepia}
\end{figure}

This section provides additional qualitative results for the inpainting task on DAVIS 2017 training/validation videos. In Figure~\ref{fig:flickering-structure-supp}, we depict additional examples where HyperCon reduces flickering and boundary distortion effects, as well as produces more realistic texture, compared to the baselines. Next, in Figure~\ref{fig:checkerboard}, we show examples in which the image-to-video model transfer baselines produce checkerboard artifacts and HyperCon does not. Finally, in Figure~\ref{fig:vcons-sepia}, we compare our method to Cxtattn-vcons in cases where Cxtattn-vcons fails to make a prediction that blends well with the known pixels due to the hue shift problem.

%
%
%
%

\newpage
\twocolumn

{\small
\bibliographystyle{ieee_fullname}
\bibliography{references}
}

\end{document}